\documentclass[11pt,a4paper]{article}
\usepackage[dvipsnames]{xcolor}
\usepackage[acceptedWithA, hyperref]{tacl2018v2}
\usepackage{comment}
\usepackage[disable]{todonotes}
\usepackage{times}
\usepackage{svg}
\usepackage{latexsym}
\usepackage{enumitem}

\usepackage{booktabs}
\usepackage{multirow}
\newcommand{\tabincell}[2]{\begin{tabular}{@{}#1@{}}#2\end{tabular}}  
\usepackage{microtype}
\usepackage{csquotes}
\usepackage{bm}
\usepackage{amsmath,amssymb, amsfonts,mathtools, bm}
\usepackage{soul}
\usepackage{adjustbox}
\usepackage{resizegather}
\usepackage{appendix}
\usepackage{etoolbox}
\usepackage{caption}
\usepackage{subcaption}
\usepackage{enumerate}
\usepackage{graphicx}
\usepackage{float}
\usepackage{scalerel}
\usepackage{svg}
\BeforeBeginEnvironment{appendices}{\clearpage}
\usepackage[ruled,vlined]{algorithm2e}
\usepackage{linguex}
\usepackage[utf8]{inputenc}
\usepackage[font=small,labelfont=bf]{caption}
\usepackage{cleveref}
\Crefname{ALC@unique}{Line}{Lines}
\Crefname{section}{\S}{\S\S}
\Crefname{section}{\S}{\S\S}
\Crefformat{section}{\S#2#1#3}
\Crefname{table}{Table}{Tables}
\Crefname{figure}{Fig.}{Fig.}
\Crefname{algorithm}{Alg}{Alg}
\Crefname{algorithm}{Alg}{Alg}
\Crefname{line}{line}{lines}
\Crefname{appendix}{\S\!\!}{\S\!\!}
\Crefname{thm}{Theorem}{}
\Crefname{prop}{Prop.\@}{Props.\@}
\Crefname{defin}{Definition}{Definitions}
\Crefname{lemma}{Lemma}{Lemmata}
\Crefname{cor}{Corollary}{Corollaries}
\Crefname{equation}{}{}
\Crefname{myexample}{Example}{Examples}

\newcommand{\note}[4][]{\todo[author=#2,color=#3,size=\scriptsize,fancyline,caption={},#1]{#4}}

\newcommand{\ryan}[2][]{\note[#1]{ryan}{violet!40}{#2}\xspace}

\DeclareMathOperator*{\argmax}{argmax}

\usepackage{relsize}
\usepackage{booktabs}
\usepackage{makecell}
\usepackage{relsize}
\newcommand{\defeq}[0]{\mathrel{\stackrel{\textnormal{\tiny def}}{=}}}

\newcommand{\bx}{{\mathbf{x}}}
\newcommand{\by}{{\mathbf{y}}}

\usepackage{array}
\usepackage{ragged2e}
\newcolumntype{P}[1]{>{\RaggedRight\hspace{0pt}}p{#1}}
\newcolumntype{X}[1]{>{\RaggedRight\hspace*{0pt}}p{#1}}
\usepackage{tikz}
\usetikzlibrary{backgrounds}
\usetikzlibrary{arrows,shapes}
\usetikzlibrary{tikzmark}
\usetikzlibrary{calc}
\usetikzlibrary{arrows,shapes,positioning,shadows,trees,mindmap}
\usepackage[edges]{forest}
\usetikzlibrary{arrows.meta}
\colorlet{linecol}{black!75}
\usepackage{xkcdcolors} 
\usepackage{tcolorbox}
\newcommand{\highlight}[2]{\colorbox{#1!17}{$\displaystyle #2$}}

\renewcommand{\highlight}[2]{\colorbox{#1!17}{#2}}

\definecolor{darkorange}{rgb}{1, 0.549, 0}
\colorlet{mhpurple}{Plum!80}

\usepackage{stmaryrd}
\newcommand{\den}[2]{\left\llbracket\mathrm{#1}\right\rrbracket_{#2}}
\newcommand{\coloredUlEntity}[2]{\textcolor{#1}{\textbf{\underline{#2}}}}
\newcommand{\coloredEntity}[2]{\textcolor{#1}{\textbf{#2}}}
\newcommand{\grayEntity}[1]{\textcolor{gray}{\textbf{#1}}}
\newcommand{\defn}[1]{\textbf{#1}}
\newcommand{\bCf}{\boldsymbol{C}_{\scaleto{\!\mathit{f}}{5pt}}}
\newcommand{\bCb}{\boldsymbol{C}_{\scaleto{\!\mathit{b}}{5pt}}}
\newcommand{\Cb}{C_{\scaleto{\!\mathit{b}}{5pt}}}
\newcommand{\Cp}{C_{\scaleto{\!\mathit{p}}{5pt}}}
\newcommand{\Ch}{C_{\scaleto{\!\mathit{h}}{5pt}}}

\newcommand{\Unprev}{U_{n-1}}
\newcommand{\mentions}{\mathcal{M}}
\newcommand{\entities}{\mathcal{E}}
\newcommand{\discourse}{\mathfrak{D}}
\newcommand{\corpus}{\mathfrak{C}}
\newcommand{\vocab}{\mathcal{V}}
\newcommand{\R}{\mathbb{R}}
\newcommand{\Un}{U_n}
\newcommand{\invf}{f_{\Un}^{\scaleto{-1}{5pt}}}

\newcommand{\weight}{\mathsf{weight}}

\newcommand{\gate}{\mathsf{gate}}
\newcommand{\forget}{\mathsf{forget}}

\newcommand{\semiring}{{ \mathcal{W}}}
\newcommand{\semiringtuple}{{\left(\semiringset, \oplus, \otimes, \zero, \one \right)}}
\newcommand{\semiringset}{{ A}}
\newcommand{\zero}{{\textbf{0}}}
\newcommand{\one}{{\textbf{1}}}

\newcommand{\Continue}{\textsc{Continue}}
\newcommand{\Retain}{\textsc{Retain}}
\newcommand{\SmoothShift}{\textsc{SmoothShift}}
\newcommand{\RoughShift}{\textsc{RoughShift}}
\newcommand{\grammaticalRole}{\textsc{grammaticalRole}}
\newcommand{\semanticRole}{\textsc{semanticRole}}
\newcommand{\Utterance}{\textsc{utterance}}
\newcommand{\PreviousUtterance}{\textsc{previous utterance}}
\newcommand{\cfCandidate}{\textsc{cf candidate}}
\newcommand{\MappingFunc}{\textsc{mention entity mapping} $f$}
\newcommand{\WeightingFunc}{\textsc{weight}}

\newcommand{\clusterOnly}{\textsc{clusterOnly}}
\newcommand{\includeSingleton}{\textsc{includeSingleton}}

\newcommand{\spanbertModel}{\textsc{spanbert}}
\newcommand{\gloveModel}{\textsc{glove}}
\newcommand{\oneHotModel}{\textsc{onehot}}

\newcommand{\ctof}{\textsc{c2f}}
\newcommand{\incremental}{\textsc{incremental}}
\newcommand{\corefqa}{\textsc{corefqa}}
\newcommand{\weak}{\textsc{weak}}
\newcommand{\strong}{\textsc{strong}}

\usepackage{stmaryrd}

\newcommand{\coloredUl}[2]{\textcolor{#1}{\underline{#2}}}
\newcommand{\redUl}[1]{\coloredUl{red}{#1}}
\newcommand{\blueUl}[1]{\coloredUl{blue}{#1}}
\newcommand{\redUlBf}[1]{\coloredUlEntity{red}{#1}}
\newcommand{\blueUlBf}[1]{\coloredUlEntity{blue}{#1}}

\title{Investigating the Role of Centering Theory\\ in the Context of Neural Coreference Resolution Systems}

\author{
Yuchen Eleanor Jiang~\;~Ryan Cotterell~\;~Mrinmaya Sachan \vspace{5pt}\\
   {\includegraphics[width=6em,height=0.8em]{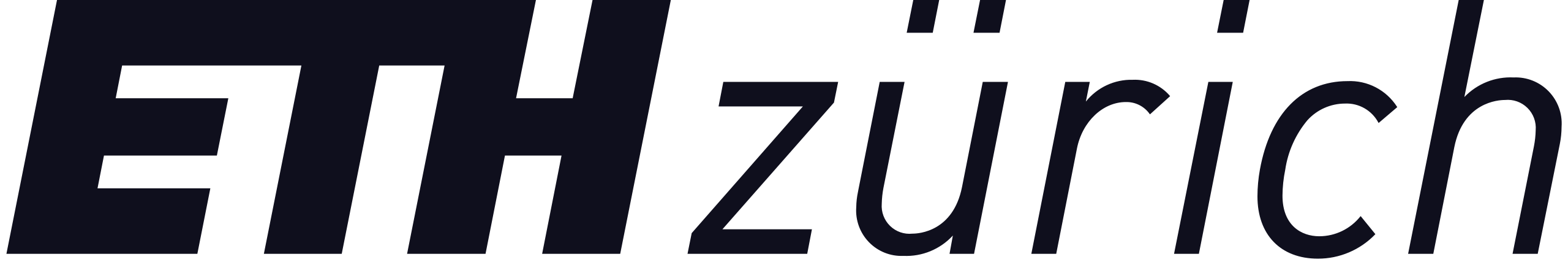}}\\ 
    \texttt{\{\href{mailto:yuchen.jiang@inf.ethz.ch}{yuchen.jiang},\href{mailto:mrinmaya.sachan@inf.ethz.ch}{mrinmaya.sachan},\href{mailto:ryan.cotterell@inf.ethz.ch}{ryan.cotterell}\}@inf.ethz.ch }
}

\date{}

\begin{document}
\maketitle

\begin{abstract}
Centering theory \cite[CT;][]{grosz-etal-1995-centering} provides a linguistic analysis of the structure of discourse. According to the theory, local coherence of discourse arises from the manner and extent to which successive utterances make reference to the same entities. In this paper, we investigate the connection between centering theory and modern coreference resolution systems. We provide an operationalization of centering and systematically investigate if neural coreference resolvers adhere to the rules of centering theory by defining various discourse metrics and developing a search-based methodology. Our information-theoretic analysis reveals a positive dependence between coreference and centering; but also shows that high-quality neural coreference resolvers may not benefit much from explicitly modeling centering ideas. Our analysis further shows that contextualized embeddings contain much of the coherence information, which helps explain why CT can only provide little gains to modern neural coreference resolvers which make use of pretrained representations.
Finally, we discuss factors that contribute to coreference which are not modeled by CT such as world knowledge and recency bias. We formulate a version of CT that also models recency and show that it captures coreference information better compared to vanilla CT.

\noindent {\includegraphics[width=1.25em,height=1.25em]{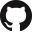}\hspace{.75em}\parbox{\dimexpr\linewidth-2\fboxsep-2\fboxrule}{\url{https://github.com/EleanorJiang/ct-coref}}}
\end{abstract}

\section{Introduction}
Centering theory \citep[CT;][]{grosz-etal-1995-centering} is a well-known theory of discourse that provides an account of the coherence of a piece of text through the manner in which successive utterances refer to the same discourse entity.
CT has served as a theoretical foundation for many NLP applications such as coreference resolution, machine translation, text generation and summarization. 
Among them, CT has been most well-studied in the context of coreference, a task of linking referring expressions to the entity they refer to in the text \citep{sidner1979towards, brennan-etal-1987-centering, iida2003incorporating, beaver2004optimization, kong-etal-2009-employing, kehler2013probabilistic}. 

Previous work has shown that there are deep connections between CT and coreference.
Referring expressions often show preference to certain
linguistic forms to indicate a reference relation to their antecedents.
For example, pronouns are often used to refer to preceding named entities, but reintroduction of the named entity leads to use of their nominal form.
These referring expressions thereby can be seen to connect the various utterances in the text and
contribute to the coherence of the overall text. Thus, it has long been believed that coherence can impose constraints on referential accessibility. See \cref{fig: coref_coherence_example} for an example.
\begin{figure}
\begin{adjustbox}{width=0.47\textwidth}
{
\centering
\begin{tabular}{ll}
\toprule[2pt]
$U_1$ & {\footnotesize \redUlBf{John} has been having a lot of \ul{trouble} arranging \redUl{his} \ul{vacation}}. \\

$U_2$ &{\footnotesize \redUl{He} cannot find anyone to take over \redUl{his} \ul{responsibilities}.} \\ 
$U_3$ &{\footnotesize \redUl{He} called up \blueUlBf{Mike} yesterday to work out a \ul{plan}.}  \\
\midrule[1pt]
$U_4$ &{\footnotesize \blueUlBf{Mike} has annoyed \redUlBf{John} a lot recently.} \\
$U'_4$ &{\footnotesize \blueUl{He} has annoyed \redUlBf{John} a lot recently.} \\
\midrule[1pt]
$U_5$ &{\footnotesize \blueUl{He} called \redUlBf{John} at 5 AM on Friday last week.} \\
\bottomrule[2pt]
\end{tabular}
}
\end{adjustbox}
\caption{An example of anaphora resolution and coherence. Mentions spans are underlined, and colors represent entity clusters. An account of coherence is closely related to the question: Why $U_4$ is better than $U'_4$? \looseness=-1
}
\vspace{-8pt}
\label{fig: coref_coherence_example}
\end{figure}

Old coreference resolution models indeed exploited this connection 
\citep{brennan-etal-1987-centering, sidner1979towards, iida2003incorporating, beaver2004optimization, kong-etal-2009-employing},
arguing that the constraints proposed by CT can serve as a useful guide for coreference resolution models~\citep{elango2005coreference, deemter2000coreferring, chai-strube-2022-incorporating}. 
However, modern coreference systems are primarily based on neural networks and are trained end to end without any explicit linguistic bias. 
A natural question is, then, \textit{whether these neural coreference resolvers work in a similar way as CT suggests and, more practically, if CT may be a useful inductive bias for neural coreference resolution systems}. 

In this paper, we attempt to provide an answer to these questions through a careful analysis of neural coreference models using various discourse metrics (referred to as centering metrics) and conducting several statistical tests. 
Because CT, at its core is a linguistic theory, and not a computational one, we first provide a computational operationalization of CT that we can directly implement (\cref{ct}).
Our operationalization requires us to concretely specify the linguistic notations present in the original work \citep{grosz-etal-1995-centering, poesio-etal-2004-centering} and 
draw conclusions about how well neural coreference resolvers accord with CT.\looseness=-1


In a series of systematic analyses (\cref{exp_res}), we first show that neural coreference resolution models achieve relatively high scores under centering metrics, indicating they do contain some information about discourse coherence, even though they are not trained by any CT signals. In addition, as shown in \Cref{fig:CT-F1}, there is a non-trivial relationship between CT and coreference, which we quantify by mutual information, between the performance of a coreference resolver and our various CT operationalizations 
~\cite{chambers1998structural, gordon1998representation}.
However, the centering scores taper off as we have more accurate coreference models (i.e., models with higher CoNLL F1):
the dependence between CT and coreference performance decreases when CoNLL F1 reaches above 50\%. This interval, unfortunately, is where all modern coreference resolution models lie. This indicates that entity coherence information is no longer helpful in improving current neural coreference resolution systems.

Next, we turn to answering the question: \textit{Where in their architecture do neural coreference systems capture this CT information?} 
Our experiments on the well-known \ctof{} coreference model with SpanBERT embeddings \cite{joshi-etal-2020-spanbert} (\cref{ana1})
 reveal that the contextualized SpanBERT embeddings
contain much of the coherence information, which explains why incorporating elements of CT only yields minor improvements to a neural coreference systems.\looseness=-1 

Finally, we explore \textit{what information required in coreference resolution is not captured by CT?} We show that CT does not capture factors such as recency bias and world knowledge (\cref{sec:dis:coherence_modeling}) which might be required in the task of coreference resolution. In order to explore the role of recency bias
, we extend our CT formulation to account for this bias by controlling the salience of centers in the CT formulation. We show that this reformulation of CT captures coreference information better compared to vanilla CT at the same centering score level. 
We end with a summary of takeaways from our work.

\begin{table*}
\begin{adjustbox}{width=\textwidth}
{
\centering
\begin{tabular}{llllll}
\toprule[2pt]
      & Utterance; mentions (elements of $\mentions$) are underlined  & $\bCf$ & $\Cp$ & $\Cb$ & Transition\\
\midrule[1pt]
$U_1$ & {\footnotesize \ul{John} has been having a lot of \ul{trouble} arranging \ul{his} \ul{vacation}}. & \makecell[l]{$\den{John}{i}$ \\ $\den{trouble}{j}$ \\ $\den{vacation}{k}$} & $\den{John}{i}$ & $\varepsilon$ & --- \\
\midrule\vspace{.1cm}
$U_2$ &{\footnotesize \ul{He} cannot find anyone to take over \ul{his} \ul{responsibilities}.} &
\makecell[l]{$\den{John}{i}$ \\ $\den{responsibilities}{l}$} &  $\den{John}{i}$ & $\den{John}{i}$ & {\footnotesize\Continue} \\ \midrule\vspace{.1cm}
$U_3$ &{\footnotesize \ul{He} called up \ul{Mike} yesterday to work out a \ul{plan}.} &
\makecell[l]{$\den{John}{i}$ \\ $\den{Mike}{m}$ \\ $\den{plan}{n}$}& $\den{John}{i}$ & $\den{John}{i}$ &  {\footnotesize \Continue} \\
\midrule\vspace{.1cm}
$U_4$ &{\footnotesize \ul{Mike} has annoyed \ul{John} a lot recently.} & \makecell[l]{$\den{Mike}{m}$ \\ $\den{John}{i}$} & $\den{Mike}{m}$ & $\den{John}{i}$ & {\footnotesize \Retain} \\
\midrule\vspace{.1cm}
$U_5$ &{\footnotesize \ul{He} called \ul{John} at 5 AM on Friday last week.} & \makecell[l]{$\den{Mike}{m}$    \\ $\den{John}{i}$} & $\den{Mike}{m}$ & $\den{Mike}{m}$ & {\footnotesize \SmoothShift} \\
\bottomrule[2pt]
\end{tabular}
}
\end{adjustbox}
\caption{An example describing centering theory with the weighting function $w$ being \grammaticalRole. Here, $[*]_i$ denotes the entity $e_i$. For each utterance, a set of mentions $\mentions$ are detected with weights, then map to a set of entities $\bCf$ (both \ul{He} and \ul{his} map to $\den{John}{i}$ in $U_2$). We sort the entities in $\bCf$ by their weights for illustration. $\Cp$ is the most weighted element in $\bCf$ ($\den{John}{i}$ is an more important entity than $\den{responsibilities}{l}$ in $U_2$). $\Cb$ is chosen from the $\bCf$ of the previous utterance.}
\label{example}
\end{table*}

\section{Coreference and Centering Theory} \label{ct}  
In this section, we overview the necessary background on coreference and centering theory in our own notation.
We define a \defn{discourse} $\discourse = [U_1, \ldots, U_N]$ of length $N$ as a sequence of $N$ utterances, each denoted as $\Un$.
We take an \defn{utterance} $\Un$ of length $M$\ryan{Should this be $M_n$? The strings don't have to be of the same length. | eleanor: M_n makes the notations quite fussy in $\mentions\left(U_n\right) = \{m_1, m_2, \ldots\}$ ? } to be a string of tokens $t_1\cdots t_M$ where each token $t_m$ is taken from a vocabulary $\vocab$.\footnote{ 
This definition of an utterance could be understood as a textual unit as short as a clause, but it also could be understood as a textual unit as long as multiple paragraphs; we have left it intentionally open-ended and will revisit this point in \Cref{exp_setup}.}
Let $\mentions\left(U_n\right) = \{m_1, m_2, \ldots\}$ be the set of mentions in the utterance $\Un$.
A \defn{mention} is a subsequence of the tokens that comprise $U_n = t_1 \cdots t_M$. Mentions could be pronouns, repeated noun phrases, and so forth, and are often called anaphoric devices in the discourse literature.

\subsection{Coreference}
Next, let $\entities$ be the set of entities in the world. 
A \defn{coreference resolver} \ryan{Isn't this more of an entity linker? | It is. You named it ''coreference resolver`` tho. :P}$f : \mentions\left(\discourse\right) \rightarrow \entities$ implements a function from the set of mentions onto the set of entities  (henceforth also referred to as the \MappingFunc).\footnote{In general, coreference resolution includes a mention detection step. In our analysis, we assume the mentions to be given. Thus, $f$ can essentially be thought of as an implementation of the entity-linking step in coreference resolution.}
In \Cref{example}, $\den{\cdot}{}$ denotes $f(\cdot)$ for illustration, i.e., a mention, e.g., Mike, is mapped to the entity $\den{Mike}{i}$.
Here we reuse the notation $\mentions(\cdot)$, where $\mentions(\discourse) \defeq \bigcup\limits_{U_n \in \discourse} \mentions(U_n)$. 
Rule-based or feature-based coreferences resolvers~\citep{hobbs1978resolving, sidner1979towards, brennan-etal-1987-centering,kong-etal-2009-employing} resolve coreference by explicitly combining CT constraints or syntactic constraints. 
Current state-of-the-art coreference resolvers are end-to-end neural models~\citep{lee-etal-2017-end, joshi-etal-2020-spanbert, wu-etal-2020-corefqa}.

\subsection{Centering Theory} \label{sec:ct}
Centering theory (CT)
offers a theoretical explanation of local discourse structure that models the interaction of referential continuity and the salience of discourse entities in the internal organization of a text.
It was one of the first formal treatments of discourse, and remains one of the most influential.
As the name suggests, CT revolves around the notion of \defn{centering}, which is, informally, the shifting of focus from one  entity to another during the discourse.
A \defn{center} is then defined as an entity in $\entities$ that is in the focus at a certain point in the discourse.
CT describes some preferences on:
a) the nature of the shift of the center from one entity to another, and
b) linguistic properties of mentions referring to the center (e.g., mentions that attach to the center are typically subjects and are preferentially pronominalized compared to others).
We offer a more formal treatment later in the section.\looseness=-1

As an example of centering theory in action, consider the discourse given in \Cref{example}: $\discourse = \left[U_1 \ldots U_5\right]$. \looseness=-1
Now, consider replacing $U_4$ with: 
\ex.\label{ex:fourth} $U'_4:$ He has annoyed John a lot recently.

\noindent Note that the resulting discourses $\discourse = \left[U_1 \ldots, U_4, U_5\right]$ and $\discourse' = \left[U_1 \ldots, U'_4, U_5\right]$ differ only by one utterance. CT
argues that $\discourse'$ is not as felicitous as $\discourse$. This is because, in the utterance $U_3$, the discourse entity $\den{John}{i}$ is the center and not $\den{Mike}{m}$, and
given a preference
for pronominalizing the center of attention, $\den{John}{i}$ should be pronominalized as well if $\den{Mike}{m}$ is pronominalized.
We will now formally define the key notions of centering theory.

\paragraph{Weighting function over Mentions.}
Let $\weight : U_n \times \mentions(U_n) \rightarrow \R$ be a weighting function on the set of mentions in utterance $\Un$. Mentions that are assigned a higher weight are more likely to link to a center, i.e. an entity in focus, in the given context. 
For example, in $U_1$ in \Cref{example}, \ul{John} is assigned the highest weight since it is the subject of the sentence, thus, is more likely to link to the center.\footnote{It is worth noting that the original presentation of \citet{grosz-etal-1995-centering}, in contrast, specifies a ranking of the entities.} 

\paragraph{Weighting function over Entities.}
Now we turn from weighting mentions to weighting entities.
Given an utterance $\Un$, let $\invf(e)$ be the pre-image of $e \in \entities$: 
\begin{equation}
 \invf(e) = \Big\{ m \mid m \in \mentions(\Un), f(m) = e \Big\}
\end{equation}
which maps an entity $e$ back to a set of mentions it links to.
Now we may lift the weighting function of a $\weight$ to an entity by having it take the highest weight attached to the mentions that link to the entity, i.e.,
\begin{equation}
    \weight\left(\Un, e\right) = \!\!\!\!\bigoplus_{m \in f_{\Un}^{-1}(e)} \!\!\!\! \weight(U_n, m)
\end{equation}
where $\oplus$ is a generic aggregator over  mentions; obvious choices are $\oplus = \max$ or $\oplus = \sum$.

The weighting function $\weight$ is arguably the most important component of centering. Previous works assume that several factors play a role in determining the weighting. Among them, grammatical roles~\citep{gordon1993pronouns, walker1998centering, grosz-etal-1995-centering, walker-etal-1994-japanese, brennan-etal-1987-centering} and semantic roles~\cite{sidner1979towards,kong-etal-2009-employing} are two factors that have been used in previous work. 
Previous work adopts the following orderings 
to define \WeightingFunc{}, respectively:
\begin{itemize}
    \vspace{-5pt}   
    \item \grammaticalRole: Pronoun(Subject) $>$ Pronoun(Object) $>$ Subject $>$ Object $>$ Others. 
    \vspace{-5pt}
    \item \semanticRole: Pronoun(Agent) $>$ Pronoun(Patient) $>$ Agent $>$ Patient $>$ Others.
    \vspace{-5pt}
\end{itemize}
Other factors may include, for example, the utterance-level first-mention advantage, i.e. entities mentioned first in a utterance are more accessible than entities mentioned second.
One could also expect external world knowledge to play a role in defining the $\weight$.
We will revisit this in \Cref{sec:dis:coherence_modeling}.

Using our weighting function, we now define a set of forward-looking centers $\bCf(\Un)$, the preferred center $\Cp(\Un)$ and the backward-looking center $\Cb(\Un)$ as follows: 


\paragraph{The Forward-Looking Centers.}
Each utterance $U_i$ in a discourse is assigned a set of forward-looking centers.
The forward-looking centers depend only on the expressions that constitute that utterance; they do not depend on any previous utterances in the discourse.

\begin{align}
    \bCf(\Un) &\defeq \Big\{ f(m) \mid m \in \mentions(\Un) \Big\} 
\end{align}
where, as mentioned earlier, $f$ is a coreference resolver.\looseness=-1

\paragraph{The Preferred Center.} The preferred center is the most prominent discourse focus in $\Un$, e.g. $\den{John}{i}$ in $U_1$ in \Cref{example}.
\begin{align}
     \Cp(\Un) &\defeq \argmax_{e \in \bCf (\Un)} \,\, \weight\left(\Un, e\right) \label{eq:cp}
\end{align}

\paragraph{The Backward-looking Center.}

The backward-looking center, however, depends mainly on the previous utterance, i.e. the most weighted element of $\bCf (\Unprev)$ that appears in $\bCf (\Un)$. In \Cref{example}, $\den{John}{i}$ as the highest-weighted element of $\bCf (U_3)$ appears in $\bCf (U_4)$, so $\Cb (U_4)$ is $\den{John}{i}$; otherwise, we will look at the second most weighted element of $\bCf (U_3)$ and so on.

\begin{align}
     \Cb(\Un) &\defeq 
     \!\! \argmax_{e \in \bCf (\Unprev)\,\cap\,\bCf (\Un)}  \!\!\!\!\!\!\!\weight\left(\Unprev, e\right) \label{eq:cb}
\end{align}
\noindent The connection between the backward-looking center and the preferred center forms the theory's key claims about coherence. That is, texts fraught with abrupt center switches from one entity to the next are perceived to be less coherent.

The key difference between CT and coreference resolution is in the design of the decision function $f$ that maps mentions to entities. While coreference models define a ranking or a classification function, CT defines a set of rules and constraints based on the relationship between the backward-looking center and the preferred center.

\subsection{Centering Metrics}
Next, we describe some metrics which use CT to assess coherence of a piece of text. We will later use these metrics to define a coherence metric which will let us study the relationship between discourse coherence as evaluated via CT and coreference.

\paragraph{\textsc{transition}.}
CT gives a formal account of how discourse involves continuous updates to a local attentional state, which is described in \Cref{tab:transition} as \defn{transitions}. The transition relations are used to hypothesize that discourses are easier to process when successive utterances are perceived as being ``about'' a unique discourse entity. Transitions have the descending preference order of \Continue $>$ \Retain $>$ \SmoothShift $>$ \RoughShift. This rule captures the essence of the theory aiming to minimize the number of focus shifts. The transition information can be used to determine the extent to which a text conforms with, or violates, the principles of centering theory. 
Naturally, we can compare the coherence of different discourses by employing the preferences of transitions, i.e. given a set of discourses of the same length, we sort them by the number of \Continue\ transitions first, then use the number of \Retain, \SmoothShift, \RoughShift\ as tie breakers sequentially.

\begin{table}
    \scalebox{0.7}{
    \centering
    \begin{tabular}{c|c|c}
    \toprule[2pt]
         & \tabincell{c}{$\Cb(\Un)=\Cb(\Unprev)$ \\ or $\Cb(\Un)$ undefined}  & $\Cb(\Un)\neq \Cb(\Unprev)$\\
    \midrule[1pt]
         $\Cb(\Un)=\Cp(\Un)$& \Continue &  \SmoothShift\\
         $\Cb(\Un) \neq \Cp(\Un))$& \Retain & \RoughShift \\
    \bottomrule[2pt]
    \end{tabular}}
    \caption{Four Types of Transitions in CT}
    \label{tab:transition}
\end{table}
Following previous literature~\cite{poesio-etal-2004-centering,karamanis2004evaluating}, we also define four centering metrics: \textsc{$\lnot$nocb}, \textsc{coherence}, \textsc{salience} and \textsc{cheap}. 

\paragraph{\textsc{$\lnot${nocb}}.}
The \textsc{$\lnot${nocb}} constraint is a predicate that returns true if $\Cb(\Un) \neq \varepsilon$. \textsc{nocb} occurs when $\Cb(\Un)$ is undefined. 
Consider the example in \Cref{example}. If $U_5$ is changed to ``Jane called Anna at 5 AM on Friday'', we will observe a \textsc{nocb}, which implies a very sharp shift of focus since no element in $\bCf(\Unprev)$ is realized in $\Un$. 

\paragraph{\textsc{coherence}.}
The \textsc{coherence} constraint is a predicate that returns true if $\Cb(\Un)=\Cb(\Unprev)$, e.g. $U_2$ to $U_3$, since the continuity of $\Cb$ is the core concept of local coherence. 

\paragraph{\textsc{salience}.}
The \textsc{salience} constraint is a predicate that returns true if $C_b(U_n)=C_p(U_n)$, indicating the maintained focus, i.e. $\Cb$ is the most salient entity ($C_p$). For example, the \textsc{salience} constraint satisfy in $U_2$, $U_3$ and $U_5$ as $\Cb$ and $\Cp$ denote the same entity.

\paragraph{\textsc{cheap}.} 
The \textsc{cheap} constraint is a predicate that returns true if $\Cb(\Un)=\Cp(\Unprev)$. 
Qualitatively, this constraint enforces that the transition taken is the easiest, i.e. the transition that causes the least inferential load on the listener~\cite{strube-hahn-1999-functional}, e.g. $U_4$ to $U_5$. 
However, if we assign $f(\mathrm{He})$ in $U_5$ to another entity, e.g. $\den{Mary}{n}$, $\Cb (U_5)$ has to be the second most weighted entity in $\bCf (U_4)$, i.e. $\den{John}{i}$, which violates the \textsc{cheap} constraint.
We measure coherence using the ratio of the transitions that satisfy those constraints to the total amount of transitions in $\discourse$, e.g. $| \left\{ \Un \mid \Unprev \rightarrow \Un \text{ satisfy } X \right\} | / (N-1)$, where $X$ is one of the constraints.

Finally, {\bf \textsc{kp}} denotes the sum of the four metrics defined above~\cite{kibble-power-2000-integrated}.

\section{Assessing Coherence} \label{assess}
Although the aforementioned centering metrics are straightforward, 
they cannot be directly used as a measurement of coherence quality
as properties of discourse that are not related to coherence also affect these statistics. For example, discourse that contains more entities and mentions tends to have a higher rate of violations. To alleviate this issue, we adopt a search-based methodology to quantify coherence, which was first brought up in \cite{karamanis2004evaluating} for text structuring.

The key assumption behind this is that among all the possible permutations of a given discourse, the original permutation should be the most coherent one. 
We randomly sample a set of permutations from the space of possible permutations of the discourse $\discourse$ for discourses longer than 5 utterances and take the entire permutation space for short discourses.
We then rank all candidate permutations of utterances by a given centering metric $M$ mentioned above. 
The explored search space is divided into sets of permutations that score better, equal, or worse than the original discourse $\discourse$ according to the metric $M$. 
The third step is to calculate the centering score
\begin{equation}
\Ch(M, f, \discourse)= \text{Worse}(M, f, \discourse) + \frac{\text{Equal}(M, f,\discourse)}{2}
\end{equation}
where the $\text{Worse}$ (or $\text{Equal})$ function denotes the number of permutations that score lower than (or equal to) $\discourse$ according to $M$ based on the coreference resolver $f$. 
Intuitively, this function assigns 1 unit of credit if original discourse $\discourse$ has a higher score than a scrambled (possibly incoherent) discourse and 0.5 unit of credit if the scores are equal.
A higher $\Ch(M,f,\discourse)$ is indicative of a set of entity clusters $f(\discourse)$ with better coherence quality.
Finally, the coherence quality $\overline{\Ch}$
on the entire corpus $\corpus$ is summarized as the average centering score\looseness=-1
\begin{equation}
\overline{\Ch}(M,f,\corpus)=\frac{1}{|\corpus|}\sum_{\discourse \in \corpus}C_h(M,\entities, \discourse)
\end{equation}

\section{Experimental Setup}\label{exp_setup}
\subsection{Datasets}
We experiment on the OntoNotes 5.0 (CoNLL-2012 shared task) dataset~\cite{weischedel2013ontonotes, pradhan2013towards}, which is a popular benchmark dataset in coreference resolution literature. 
The dataset also contains rich syntactic and semantic annotations such as part of speech, syntax and semantic roles, which is useful in our analysis.
There are 2802, 343, and 348 documents in the training, development, and test set, respectively. 
It is worth noting that all previous work on CT has been based on much smaller datasets with shorter documents as well as fewer documents overall.
For example, the Ontonotes dataset has documents that are on average 75 times longer than the \textit{GNOME} dataset~\cite{poesio-etal-2004-centering}
used in several previous CT papers \citep{poesio-etal-2004-centering, karamanis2004evaluating}.

\subsection{Coherence Model Parameters} \label{sec:parameter}
Centering theory is an abstract theory in that it discusses focality using high-level semantic notions, e.g. the salience of an entity in an utterance.
However, to make very specific claims about centering theory using data, we have to make these high-level notions precise. 
For instance, we have to precisely define what we mean by utterance and there are multiple choices, e.g. a clause, sentence or a paragraph. 
We describe and justify our choices in the operationalization of CT below.
\paragraph{\Utterance.} The \Utterance\ parameter is set to \textsc{sentences}.
In principle, one could consider other linguistic units in future work.

\paragraph{\PreviousUtterance.} The \PreviousUtterance\ parameter can be either \textsc{true} or \textsc{false}. This parameter dictates whether we should ignore \emph{null utterances}, i.e.,  utterances that do not contain  mentions such as one-interjection-utterance ``Uh-oh...''. We set this parameter to \textsc{true}. \looseness=-1

\paragraph{\cfCandidate.} The \cfCandidate\ parameter can take on values such as \clusterOnly, \includeSingleton, which stand for only taking into account entities that mentioned more than once or including singletons, respectively.
This parameter has never been considered in previous centering literature since entities are always manually annotated in the small-scale datasets they experimented on.
However, in real-world datasets, such as Ontonotes 5.0, this parameter must be set appropriately. In our experiments, both \clusterOnly{} and  \includeSingleton{} are considered.

\paragraph{\WeightingFunc.} As mentioned in \Cref{sec:ct}, we consider \grammaticalRole{} and \semanticRole{} in our experiment. For \grammaticalRole\ weighting function, the entire OntoNotes test set is used, while the experiment with \semanticRole\ weighting function is conducted on a subset of the OntoNotes testset where semantic role annotations are available. This subset consists of 61.2\% of the documents.



\subsection{Coreference Models Parameters} \label{sec:coref_para}
We explore several coreference resolution models with various embedding approaches -- SpanBERT~\cite{joshi-etal-2020-spanbert}, GloVe~\cite{pennington2014glove}, One-hot, using different amounts of training data ($10\%, 20\%, \ldots, 100\%$ of the OntoNotes training set). Each model is trained for 40 epochs and 5 independent runs with different seeds.
The models and checkpoints we used in our experiments and analysis are:
\ctof~\cite{lee-etal-2018-higher} (with \spanbertModel-base, \spanbertModel-large, \gloveModel\ and \oneHotModel\ embeddings),
\corefqa~\cite{wu-etal-2020-corefqa}, \incremental~\cite{xia-etal-2020-incremental}.
In total, there are 8,600 checkpoints. 
For \spanbertModel-large, we set \textsc{inference\_order} to 2 with \textsc{coarse\_to\_fine} being true, while the rest of \ctof{} models adopt \textsc{inference\_order} of 1 without \textsc{coarse\_to\_fine} tuning. \oneHotModel\ models use one-hot word vectors as input and do not leverage any pretrained embeddings. For both \gloveModel\ and \oneHotModel, we add another LSTM context layer after the embedding layer as in \cite{lee-etal-2017-end} and set the depth of this LSTM layer to be 1. For all the models mentioned above, the ratio of the number of mentions to the total number of words in the document (\textsc{spans\_per\_word}) is set to 0.4, the dimension of all feed-forward networks is 1500 and the dimension of the feature fed into feed-forward networks is 20. 
The upper bound number of spans is \textsc{max\_antecedents} of 50. We also filtered out spans that are longer than 30 tokens. 
For \corefqa\ and \incremental, we follow the parameter settings specified in their original papers.


\section{Experimental Results} \label{exp_res}
We first investigate the relationship between centering and coreference. 

\subsection{Comparing a Strong and a Weak Coref System on Centering Measures} \label{sec:strong_and_weak}
To start with, we compare the centering scores of \MappingFunc{} being \textsc{gold}, \textsc{strong} and \textsc{weak} in \Cref{tab:CenteringPerformance}, where \textsc{gold} denotes a ``system''\ryan{Why is this in scare quotes?| it is not a real system.} where the ground truth annotations are used for \MappingFunc{}, \textsc{strong} denotes a well-trained \ctof-\spanbertModel-large coreference model (with a CoNLL F1 of 78.87\%), and \textsc{weak} denotes a underfitted \ctof-\spanbertModel-base model (with a CoNLL F1 of 17.51\%).\footnote{\textsc{strong} is trained for 40 epochs untill convergence, while \textsc{weak} is trained only for 2 epochs.}
Four instantiations of CT are provided and we report statistical significance using the two-tailed permutation test with $p < 0.05$.

The first observation in \Cref{tab:CenteringPerformance} is that no centering scores of the coreference system significantly exceed the upper bound provided by the ground truth annotations, regardless of the instantiations. 
It appears that ground truth annotations of reference links are more consistent with the centering theory constraints than coreference annotations, which verifies our CT operationalization.
It is also worth noting that all the centering metrics of \textsc{gold} are above 70\%. 
This, from a coherence assessment perspective~\citep{barzilay-lapata-2008-modeling, karamanis2004evaluating}, indicates that the CT information is sufficient to distinguish the original discourse from randomly sampled permutations to some extent, supporting the claims about local coherence of the centering theory, particularly the CB existence claim. 

On the other hand, despite not having been trained with any CT-related supervision, the \textsc{strong} model achieves high centering scores (even not being outperformed by \textsc{gold} on \textsc{$\lnot$nocb}, \textsc{coherence}  and \textsc{salience} when \cfCandidate{} is considered to be \clusterOnly), indicating that it does contain substantial coherence information.
In contrast, the \textsc{weak} performs significantly worse on the centering scores,  which suggests that CT can be used as a goal for optimizing coreference performance.

\begin{table}[t]
    \centering
\begin{adjustbox}{width=0.49\textwidth}
    \begin{tabular}{r|c|c|c|c|c}
\toprule[2pt]
\multirow{2}{*}{\grammaticalRole}    & \multicolumn{1}{c|}{\textsc{gold}}   & \multicolumn{2}{c|}{\strong} & \multicolumn{2}{c}{\weak} \\
 & &  CO & IS  & CO & IS\\
    \midrule[1pt]
\textsc{$\lnot$nocb} & \textbf{82.7}                 & 82.6$^\star$                 & 80.6$^\dag$                 & 46.5$^\dag$                 & 44.7$^\dag$                 \\
\textsc{cheap}       & \textbf{79.1}                 & 77.7$^\dag$                 & 74.5$^\dag$                 & 45.6$^\dag$                 & 42.1$^\dag$                 \\
\textsc{coherence}   & \textbf{76.7}                 & 76.0$^\star$                 & 75.7$^\dag$                 & 45.3$^\dag$                 & 45.1$^\dag$                 \\
\textsc{salience}    & \textbf{83.8}                 & 82.6$^\star$                 & 83.4$^\dag$                 & 46.0$^\dag$                 & 46.9$^\dag$                 \\
\textsc{kp}          & \textbf{83.6}                 & 82.6$^\dag$                 & 81.7$^\dag$                 & 46.4$^\dag$                 & 42.2 $^\dag$                \\
\textsc{transition}  & \textbf{79.3}               & 78.7$^\star$                 & 73.7$^\star$                 & 61.1$^\star$                 & 52.7$^\star$                 \\
   \bottomrule[2pt]
    \end{tabular}
\end{adjustbox}
\begin{adjustbox}{width=0.49\textwidth}
 \begin{tabular}{r|c|c|c|c|c}
    \toprule[2pt]
\multirow{2}{*}{\semanticRole}    & \multicolumn{1}{c|}{\textsc{gold}}   & \multicolumn{2}{c|}{\strong} & \multicolumn{2}{c}{\weak} \\
 & &  CO & IS  & CO & IS\\
    \midrule[1pt]
\textsc{$\lnot$nocb} & \textbf{78.9}                 & 77.9$^\dag$                 & 76.2$^\dag$                 & 42.2$^\dag$                 & 41.0$^\dag$                 \\
\textsc{cheap}       & \textbf{76.9}                 & 75.2$^\dag$                 & 72.2$^\dag$                 & 43.2$^\dag$                 & 39.5$^\dag$                 \\
\textsc{coherence}   & \textbf{71.5}                 & 70.2$^\dag$                 & 70.1$^\dag$                 & 40.4$^\dag$                 & 40.1$^\dag$                 \\
\textsc{salience}    & \textbf{79.7}                 & 77.4$^\dag$                 & 78.6$^\dag$                 & 41.0$^\dag$                 & 42.3$^\dag$                 \\
\textsc{kp}          & \textbf{79.6}                 & 78.6$^\dag$                 & 78.0$^\dag$                 & 42.2$^\dag$                & 41.5$^\dag$                 \\
\textsc{transition}  & \textbf{74.9}                 & 73.7$^\dag$                 & 74.2$^\dag$                 & 52.7$^\dag$                 & 54.2$^\dag$  \\
   \bottomrule[2pt]
    \end{tabular}
\end{adjustbox}
    \caption{The comparison of different centering metrics by a permutation-based methodology with \MappingFunc\ being gold annotations and two coreference resolvers. \textit{CO} and \textit{IS} stand for \cfCandidate\ being \clusterOnly\ and \includeSingleton, respectively. $^\dag$ represents that the current model is significantly outperformed by the corresponding right one; $^\star$ represents the opposite ($\alpha = .05$).}
    \label{tab:CenteringPerformance}
    \vspace{-0.2in}
\end{table}





\subsection{Scaling Up: Exploring the Relationship between Coreference and Coherence} \label{exp3} 
\begin{figure}[t]
\centering
\includegraphics[width=0.49\textwidth,trim={13cm 1.5cm 5cm 1.6cm} ,clip]{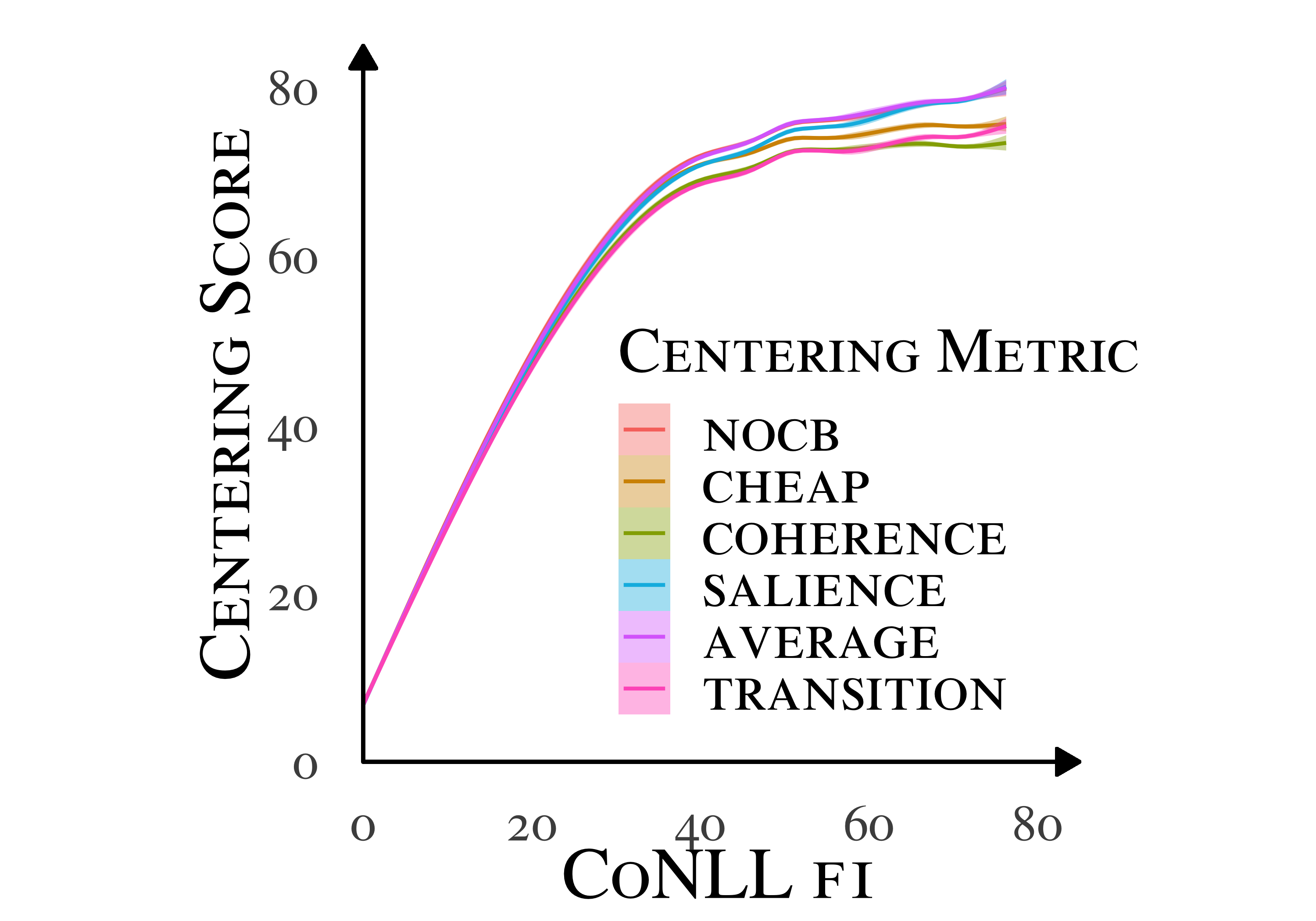}
\caption{Centering scores as a function of coreference performance (measured by CoNLL F1). }
\vspace{-0.1in}
\label{fig:CT-F1}
\end{figure}

To further systematically investigate the shape of the relationship between CT and coreference, we scale up our experiment. 
We compute the centering scores and the CoNLL F1 scores of all 8,600 checkpoints listed in \Cref{sec:coref_para}, with the weighting function being \grammaticalRole{} and \cfCandidate\ being \clusterOnly{}. 
As shown in  \Cref{fig:CT-F1}, there is a monotonic relationship between the centering scores and the CoNLL F1. The centering metrics have high positive correlations ($.883$) and high mutual information ($.451$) with CoNLL F1, with significance tested by the \textit{t}-test ($p$-value $<.01$). 
As the CoNLL F1 increases, the curve is flattened out, and the centering scores stop growing when the CoNLL F1 reaches a certain level (about 50\%). 
This suggests that the utilization of CT is more useful for improving coreference models; however, it provides little information gain when the predicted coreference chains are already of high quality.
We will explore the mismatch between the CT objective and the coreference objective later in \Cref{sec:dis:coherence_modeling}.

\subsection{Where do Coreference Models Capture Centering Information?} \label{ana1}
As shown in \Cref{exp3}, unlike feature-based or rule-based systems which explicitly leverage discourse-level coherence information by combining CT rules, the usefulness of CT for coreference is not guaranteed for modern end-to-end neural coreference systems. 
These neural models get surprisingly high coreference performance even though they are trained without any coherence signals.
A natural question to ask here is \emph{why}, i.e., which components of the neural models account for coherence modeling?
Therefore, in this section, we will be concerned with answering this scientific question; in particular, two components of the neural models, namely the embedding layer (i.e. pretrained LMs) and the coreference resolver (i.e. coreference training), will be investigated and compared.
We analyze these two possible sources of coherence information using 6,000 \ctof{} model checkpoints. For simplicity, we refer the \ctof\ model with SpanBERT-base, GloVe and one-hot embeddings as \spanbertModel, \gloveModel\ and \oneHotModel{}, respectively.





\begin{figure}
\centering
\includegraphics[width=0.49\textwidth,trim={8cm 0.5cm 6cm 0cm},clip]{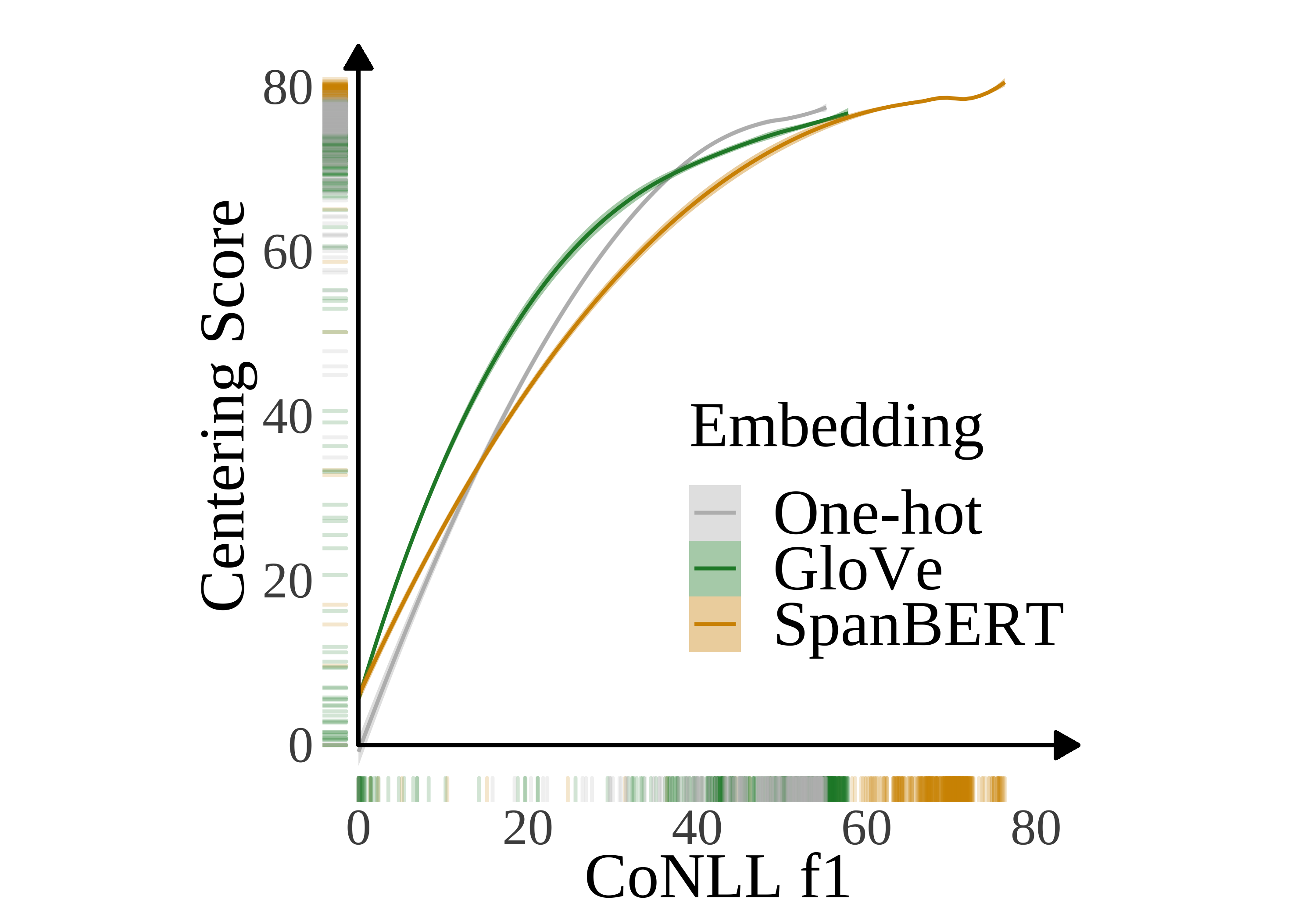}
\caption{The centering score \textsc{kp}  a function of CoNLL F1 for \spanbertModel, \gloveModel\ and \oneHotModel\ models. The x-ticks and y-ticks represent the distribution of CoNLL F1 and \textsc{kp} of the checkpoints.}
\label{fig:Glove-vs-Spanbert-all}
\end{figure}

\paragraph{Centering scores vs. CoNLL F1.}
Our first objective is to investigate how the performance of coreference and coherence varies as a function of the embedding layer. 
A version of \Cref{fig:CT-F1} where the models are grouped by the types of their embedding layers is shown in \Cref{fig:Glove-vs-Spanbert-all}.
We observe that the centering scores of \spanbertModel{} correlate most closely with its CoNLL F1 scores, that is, \spanbertModel{} achieves greater gains on coherence modeling when training with the coreference objective.
This indicates that the coherence information contained in the pretrained contextualized embedding is the most useful for coreference resolution.

\begin{figure}[t]
\centering
\includegraphics[width=0.49\textwidth,trim={4.5cm 11cm 0cm 6cm},clip]{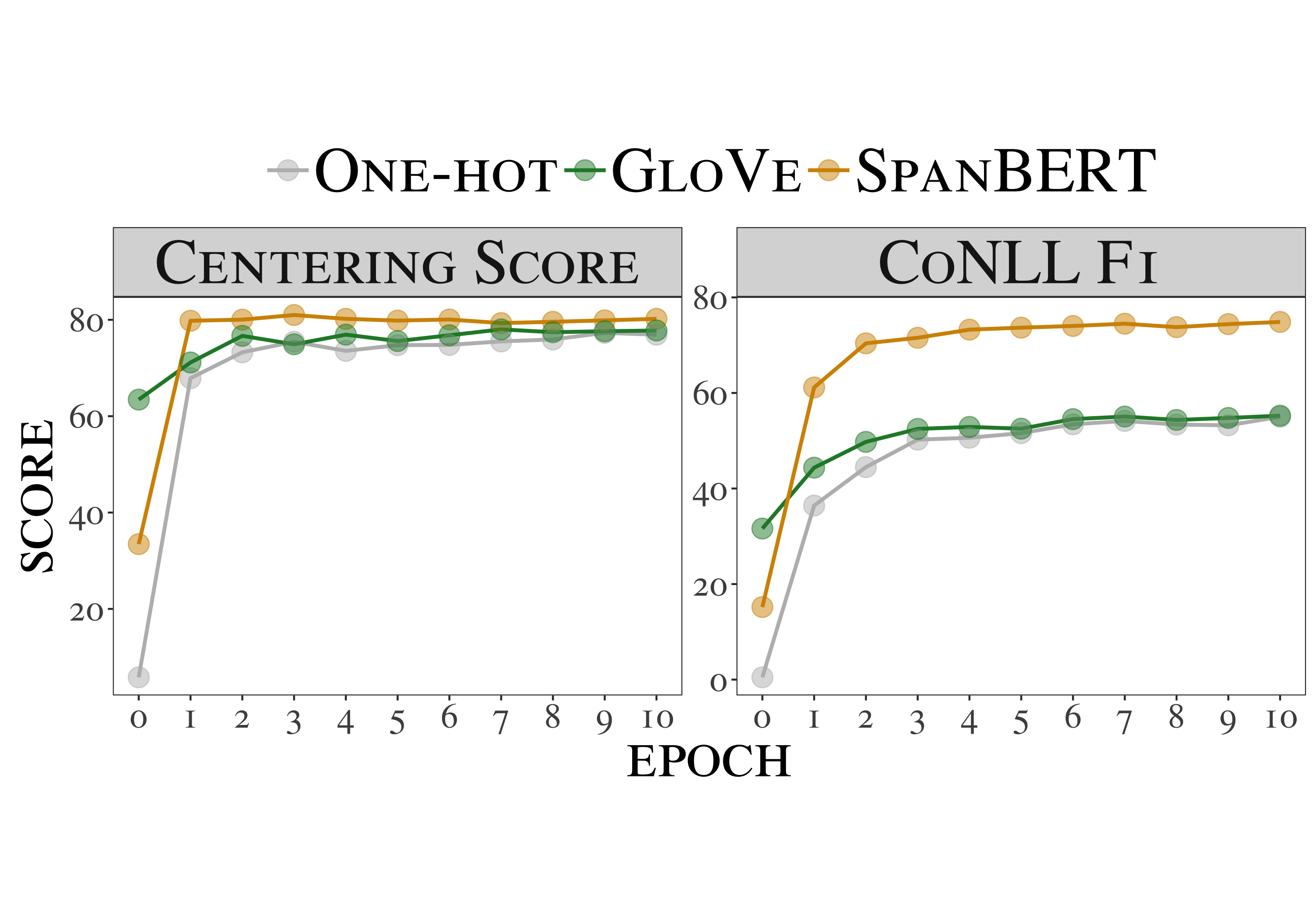}
\caption{CoNLL F1 score (square: y-axis on the left) and the centering score \textsc{kp}  (triangle: y-axis on the right) as a function of training epoch for \spanbertModel, \gloveModel\ and \oneHotModel. Models are trained on 100\% of the OntoNotes training data. }
\label{fig:Glove-vs-Spanbert-epoch}
\end{figure}

\paragraph{Training epoch.}
We further show both the CoNLL F1 and the centering scores as a function of training epoch in \Cref{fig:Glove-vs-Spanbert-epoch}. For each epoch, we calculate the mean of five independent runs.
While the centering scores of \oneHotModel{} and \gloveModel{} increase smoothly over time, \spanbertModel{} reaches a relatively high level of centering score at a very early training epoch and does not improve thereafter.
This result suggests that \spanbertModel{} contains more coherence information than \oneHotModel{} and \gloveModel{}.
In addition, the centering score for \oneHotModel{} is initially higher than that of \gloveModel{}. 
One hypothesis for this observation is that the sparse one hot encoding enables \oneHotModel{} to rapidly learn coherence information from Ontonotes in the early stages of the training process.

\begin{figure}[t]
\centering
\includegraphics[width=0.49\textwidth,trim={4cm 5cm .2cm 2cm},clip]{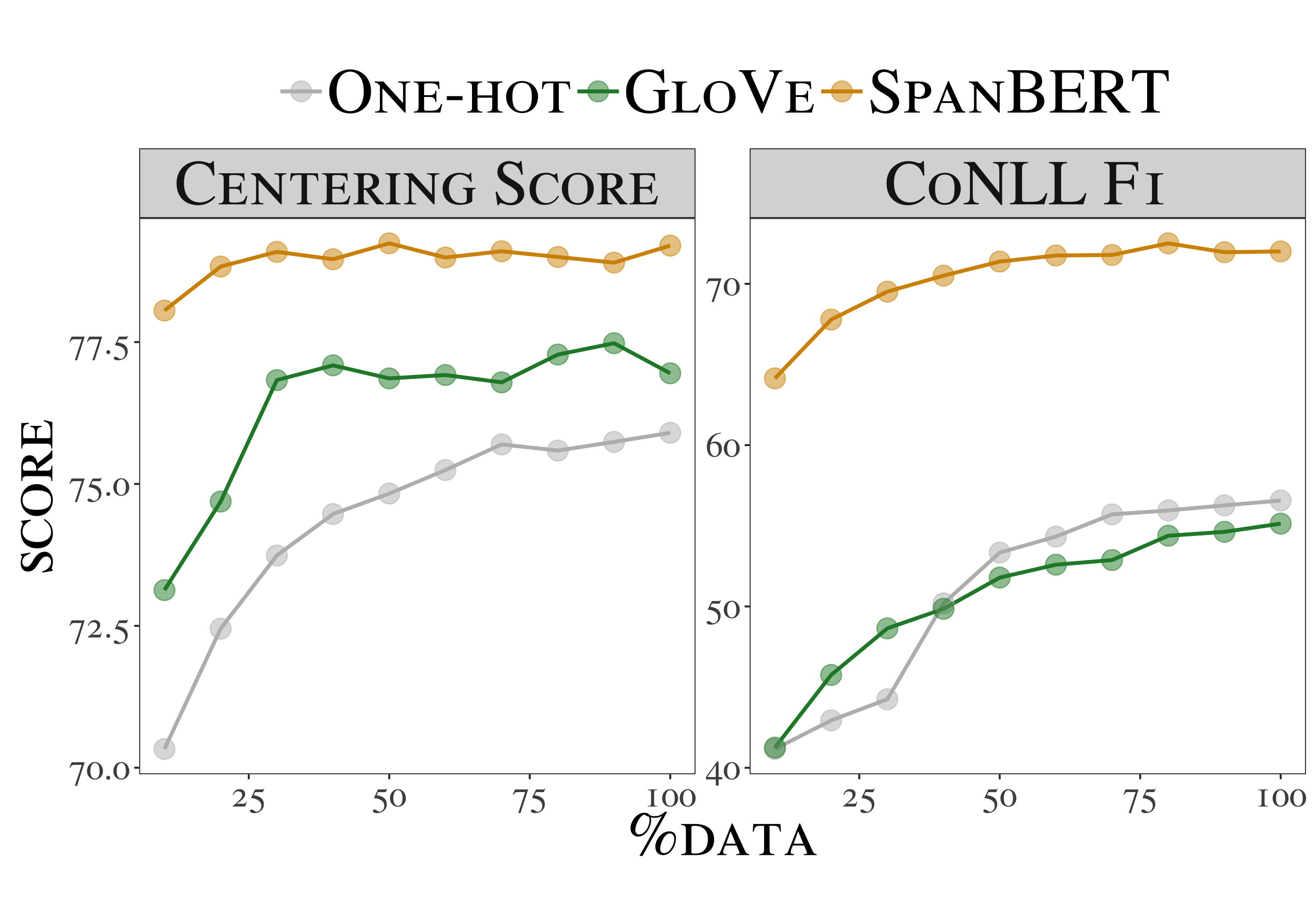}
\caption{CoNLL F1 score and the centering score \textsc{kp} as a function of the amount of coreference training data for \spanbertModel, \gloveModel\ and \oneHotModel. X-axis represents the ratio of the amount of data used in training to the original size of the Ontonotes training dataset.}
\label{fig:Glove-vs-Spanbert-training-data}
\end{figure}

\paragraph{Amount of Training Data.}
\Cref{fig:Glove-vs-Spanbert-training-data} shows the CoNLL F1 and centering scores as a function of training data. The best performing models across different training epochs are considered, resulting in 150 checkpoints.
As shown in \Cref{fig:Glove-vs-Spanbert-training-data}, the centering scores and the CoNLL F1 increase as the amount of coreference training data increases, suggesting that the practice of coreference training does lead to improvements in coherence modeling.
There is, however, a smaller growth rate for \spanbertModel{} compared to \gloveModel{} and \oneHotModel{}. It indicates that the coherence information contained in the coreference training may overlap substantially with that contained in the pretrained embeddings.

\begin{figure}
   \includegraphics[width=0.48\textwidth,trim={0cm 0cm 0cm 0cm} ,clip]{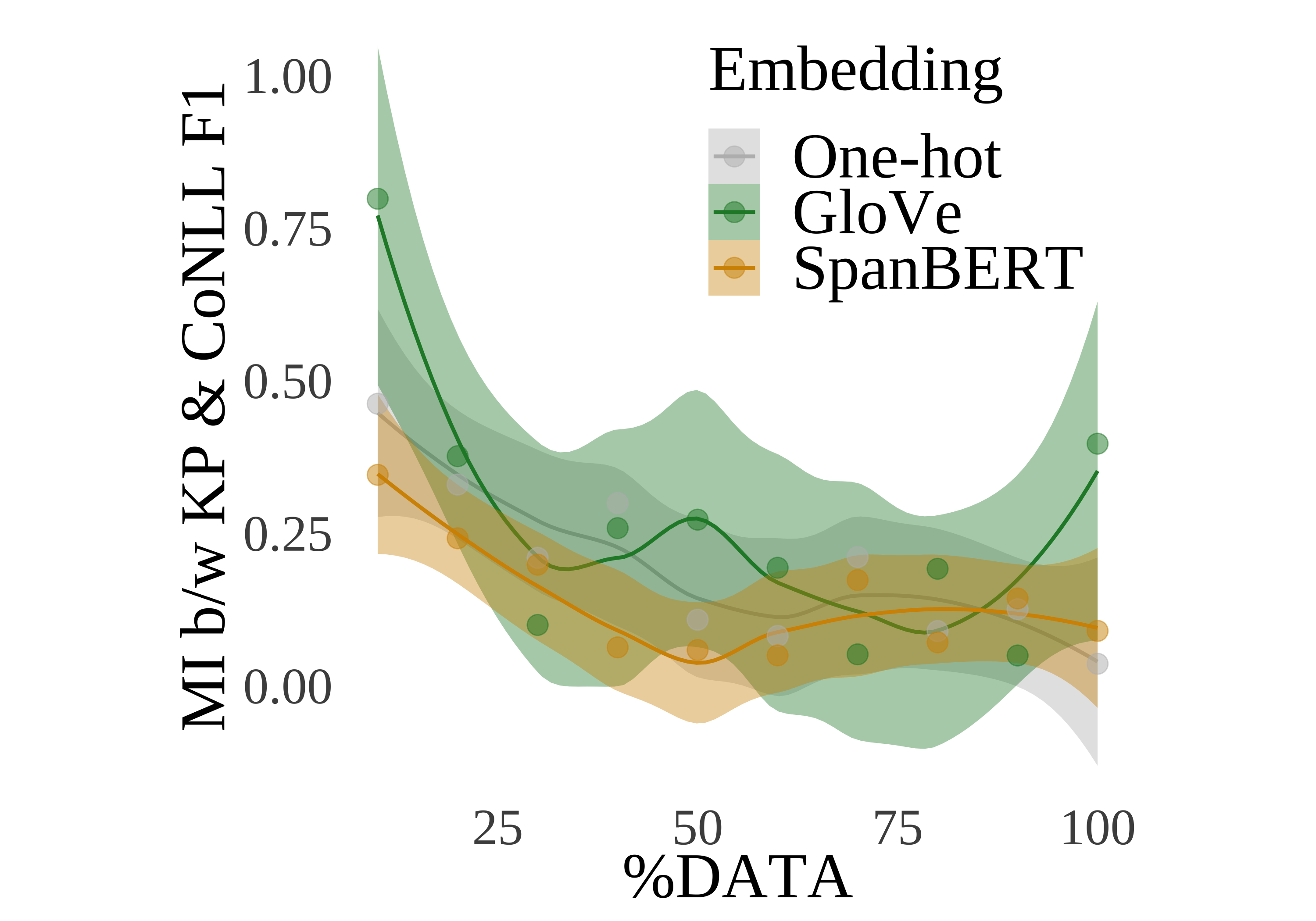}
\caption{The mutual information between CoNLL F1 and the centering score \textsc{kp} as a function of the amount of coreference training data for \spanbertModel, \gloveModel\ and \oneHotModel{}, respectively. }
\vspace{-0.2in}
\label{fig:MI-data}
\end{figure}

\paragraph{Mutual Information.}
So far, we have shown 1) pretrained contextualized embeddings contain a substantial amount of coherence information that can be utilized for coreference resolution; 2) coreference training further improves the model's ability to perform coherence modeling.
In order to compare the contributions of the embedding layer and coreference training to coherence modeling, we quantify the relationship between coherence and coreference by mutual information (MI).
We report the MI between the centering scores and CoNLL F1 for different embeddings in \Cref{fig:MI-data}.\footnote{
All 6,000 checkpoints are used for MI estimation. We adopt the default bin estimation in \texttt{infotheo} package in R for mutual information estimation. The same trends were observed when we tested other MI estimation techniques, including KDE, KSG, and partitioning.}

The MI between the centering scores and CoNLL F1 is smaller for \spanbertModel{}, especially when trained with fewer coreference annotations, suggesting that implementing the pretrained LM as the embedding layer already provides sufficient information about coherence modeling that overlaps with coreference. Therefore, the model learns mainly information that does not overlap with centering-based coherence modeling from coreference annotations. The ineffectiveness of CT for modern neural coreference systems can be attributed to this finding.
This also suggests that coreference resolution needs more than coherence modeling.


\begin{table}
\centering
\resizebox{\columnwidth}{!}{
\begin{tabular}{ll}
\hline
1.a) &To express  \coloredUlEntity{blue}{its} determination ,  \coloredUlEntity{blue}{the Chinese securities} \\ 
&\coloredUlEntity{blue}{regulatory department} compares \coloredEntity{red}{this stock reform} \\
&to \grayEntity{a die that has been cast}.  \\
1.b) &It takes time to prove whether \textcolor{red}{\textbf{\underline{the stock reform}}} can \\
& really meet expectations, and whether any deviati...  \\
1.c) & \textcolor{cyan}{\textbf{Dear viewers}} , \textcolor{magenta}{\textbf{\underline{the China News}}} will end here.  \\
1.d) &Thank \textcolor{cyan}{\textbf{\underline{you}}} \textcolor{cyan}{\textbf{\underline{everyone}}} for watching . \\
1.e) &Coming up is \textcolor{darkorange}{\textbf{\underline{the Focus Today}}} hosted by \textcolor{gray}{\textbf{Shilin}}. \\
1.f) &Good-bye , \textcolor{cyan}{\textbf{\underline{dear viewers}}}. \\
\hline
2.a) & Among unanswered questions is what \coloredEntity{blue}{PASOK's} cut \\ 
& was from the \$210 million \coloredUlEntity{red}{Mr. Koskotas} pinched. \\
2.b) & \coloredUlEntity{green}{Two former ministers} were so heavily implicated in \\
& the \coloredEntity{red}{Koskotas} affair that \coloredEntity{blue}{PASOK} members of Parlia-\\
& ment voted to refer \coloredUlEntity{green}{them} to the special court. \\
2.c) & But ... millions of drachmas \coloredUlEntity{red}{Mr. Koskotas} funneled \\
& into \grayEntity{New Democracy} coffers. \\
\hline
\end{tabular}
}
    \caption{One linguistic phenomena in coreference resolution that CT does not treat: recency of mentions. $\Cp$'s are \underline{underlined}. Singletons are marked in \textcolor{gray}{\textbf{gray}}. }
    \label{tab:example}
\end{table}



\section{What's needed for Coreference and not captured by CT?} \label{sec:dis:coherence_modeling}
As stated earlier, CT is a theory of coherence of a piece of discourse. So far, we have argued why CT should be expected to impinge on the coreference resolution task. At the same time, we have also shown that the the inductive bias introduced by CT is not enough for coreference resolution. A natural question to ask then is, are there other linguistic factors that contribute to coreference which are not captured by CT? If so, what are those factors?

In this section, we attempt to answer: \textit{What are the factors that contribute to coreference resolution which are not captured by CT?}
In particular, we examine the predictions made by our operationalization of CT 
and study various linguistic phenomena, such as recency of mentions, discourse relations and world knowledge.

\subsection{Modeling Recency Bias}
One aspect of anaphora that CT does not treat directly is recency of mentions, i.e., entities mentioned most recently are more accessible than entities mentioned earlier ~\cite{gernsbacher-1989-building}.
Inspired by the structure building framework proposed by \citet{gernsbacher-1990-language}\footnote{In this framework, language comprehension can be characterized as structure building: they maintain what they refer to as \emph{memory nodes}, which are, in essence, also entities but have a slightly broad definition than the forward-looking centers defined in the original CT framework -- they are not required to be realized in the current utterance. In other words, memory nodes are \emph{global} throughout the text while forward-looking centers in CT are \emph{local} --  the entities in $\bCf(U_{n-1})$ that are not realized in $U_n$ are removed from the set and will be treated as new entities if they ever get a chance to appear again in $U_{n+1}$.}, 
we provide a more refined account of CT that can account for a broader range of linguistic phenomena, particularly the principle of recency.
Formally, we expand the notion of the backward-looking center in \Cref{sec:ct} from a binary distinction to a scalar one.
\paragraph{Backward-looking Centers.}
Given a semiring $\semiring = \semiringtuple$, 
the backward-looking centers $\bCb$ are now defined as a set of weighted entities over $\semiring$.
$\bCb$ could now be updated recursively throughout the entire discourse: 
\begin{align} \label{eq:bcb}
     \tikzmarknode{a}{\highlight{red}{$\bCb(\Un)$}} &\defeq \tikzmarknode{b}{\highlight{red}{$\bCb(\Unprev)$}} \cup \tikzmarknode{c}{\highlight{darkorange}{$\bCf(\Unprev)$}}
\end{align}
In words, the backward-looking centers of $\Un$ are now defined as the union of the existing entities in comprehenders' mental latent representation and the newly appeared entities in $\Un$.
The weights of $\bCb(\Un)$ are defined recursively as:
\vspace{6pt}
\resizebox{\linewidth}{!}{
\begin{minipage}{\linewidth}
\begin{align}
\label{eq:bcbw}
     &\tikzmarknode{ts}{\highlight{red}{$\semiring(\bCb(\Un))$}}  \defeq \forget\left( \tikzmarknode{td}{\highlight{red}{$\semiring(\bCb(\Unprev)))$}} \right) \oplus \\  
    & \,  \gate\left(\tikzmarknode{s}{\highlight{blue}{$\bCf(\Un), \bCf(\Unprev)$}} \right) \otimes \tikzmarknode{y}{\highlight{darkorange}{$\weight\left(\Unprev, \bCf(\Unprev)\right)$}} \nonumber 
\vspace{\baselineskip}
\end{align}
\begin{tikzpicture}[overlay,remember picture,>=stealth,nodes={align=left,inner ysep=1pt},<-]
    \path (ts.north) ++ (-3em, 0) node[anchor=south west,color=red!67] (scalep){\textbf{discourse-level accessibility of centers}};
    \path (s.south) ++ (0,0) node[anchor=north, color=blue!67] (mean){\textbf{interaction b/w $\Un$,$\Unprev$}};
    \path (y.south) ++ (0,0) node[anchor=north,color=darkorange!67] (mean){\textbf{utterance-level salience}};
\end{tikzpicture}
\end{minipage}
}
\vspace{2pt}
where $\forget(\cdot)$ is a forget function and $\gate(\cdot, \cdot)$ is a gating function.
In \Cref{eq:bcb} and \Cref{eq:bcbw}, $\bCb$ could be understood, in \citet{gernsbacher-1990-language}'s structure building framework, as a set of memory nodes, and $\bCf$ could be seen as the incoming stimuli that activates, enhances or suppresses those memory nodes by reweighing $\bCb$.

It is worth noting that $\semiring$ mediates the accessibility of centers at the discourse level and thus differs from the weighting function $\weight$ that mediates the salience of centers at the utterance level. 
If we take a closer look at \Cref{eq:bcbw}, we see that \Cref{eq:cb} is just a special case of it, where $\semiring = \{\R, +, \times, 0, 1 \}$, $\forget(x) = 0$ and $\gate(\bx, \by) = \bx \wedge \by$. 
Note that a zero constant forget function  $\forget$ indicates that the original CT framework does not take into account recency.\footnote{To mitigate this issue, \citet{walker1998centering} integrates centering with a cache model of attentional state, which is limited to 3 sentences, can be viewed as the forget window being 3.}

$\Cb$ is now defined as  the most highly ranked entity in $\bCb$:

\begin{align}
     \Cb(\Un) &\defeq \argmax_{e \in \bCb (\Un)} \,\, \semiring\left(e \right) \label{eq:cp}
\end{align}

\begin{table}[t]
    \centering \small
 \begin{tabular}{c|c|c}
    \toprule[2pt]
\textbf{Metric} & Cor. & MI\\
    \midrule[1pt]
    \textsc{kp} & .883 & .451 \\
    \textsc{kp}-recency & .912(+.029)$^\dag$ & .632 (+.181)$^\dag$ \\
   \bottomrule[2pt]
    \end{tabular}
    \caption{The correlation and MI between the centering metric \textsc{kp} and CoNLL F1 of both the original CT operationalizion and the refined operationalizion with the trainable forget gate. $^\dag$ denotes $p < 0.05$. }
    \label{tab:recency}
\end{table}

\paragraph{Opertionalization.}
To provide an operationalizion of the principle of recency, we adopt a one layer feed-forward neural network to be a trainable forget function $\forget(\bx)$ and discard the gate function, i.e., $\gate(\bx, \by) = \one$.\footnote{It is worth noting that in addition to recency, this refined framework could also take into account more discourse-level linguistic preferences that the original CT could not capture. For example, the advantage of the discourse-wise first-mentioned entities (referred to as \textit{first-mention bias}) could be taken into account by assignment a small decay factor $\gamma$'s to those entities in the forget function. Another example is \textit{grammatical role parallelism bias}, which can only be accounted for by taking into account the interaction between the syntactic structures of two adjacent utterances. The exploration of other linguistic phenomena with this refined framework is out of the scope of this work and left to future work.} 
\Cref{tab:recency} shows the correlation and mutual information bewteen the centering score \textsc{kp} and CoNLL F1. The same setting as in \Cref{fig:MI-data} in \Cref{ana1} is adopted.
Fisher's Z transformation and the corresponding test are conducted for the comparison of correlation coefficients and mutual information.
The refined operationalization with the trainable forget gate yields both a higher correlation and a higher MI, indicating that this variation captures more aspects of coreference in its coherence model.

\begin{table}
\centering
\resizebox{\columnwidth}{!}{
\begin{tabular}{p{40pt}p{160pt}}
\toprule[2pt]
\textsc{result} & \coloredEntity{blue}{Mitt} flew to San Diego this weekend. \\
& \coloredEntity{blue}{He} was therefore able to visit several campaign donors. \\ 
\midrule[1pt]
\textsc{parallel} & \coloredEntity{blue}{Mitt} flew to San Diego this weekend. \\
& \coloredEntity{red}{Rick} stayed in Kansas to campaign. \\
\bottomrule[2pt]
\end{tabular}
}
    \caption{Different discourse relations have different demands on coreference for coherence modeling. }
    \label{tab:discourse_relations}
\end{table}

\begin{table}
\centering
\small
\resizebox{\columnwidth}{!}{
\begin{tabular}{p{1pt}p{200pt}}
\toprule[2pt]
a) & \coloredEntity{blue}{Ukrainian} President \textcolor{pink}{Volodymyr Zelensky} on Monday delivered a virtual speech to world leaders attending \textcolor{gray}{the World Economic Forum in Davos, Switzerland}, urging \textcolor{gray}{them} to impose "maximum" sanctions on \coloredEntity{red}{Russia} for invading \textcolor{pink}{his} country.\\
b) & \textcolor{pink}{Zelensky} said more nations should embargo \coloredEntity{red}{its} oil and block \coloredEntity{red}{its} banks.  \\
\bottomrule[2pt]
\end{tabular}
}
    \caption{An example where world knowledge must be taken into account for coreference. The correct choice of antecedent for \coloredEntity{red}{its} requires the understanding of \textit{sanction} and \textit{invading}.}
    \label{tab:world_knowledge}

\end{table}

\subsection{Discourse Relations and World Knowledge}
CT, along with our refined framework, only provide accounts for \emph{attention}. However, discourse relations and world knowledge can also be important for modelling anaphora. 
\Cref{tab:discourse_relations} illustrates via an example the influence that different discourse relations might have on the relationship between coherence and coreference.  While two mentions being coreferent is crucial for the example of \textsc{result} to be coherent, the coherence of the example of \textsc{parallel} has nothing to do with whether the center stays unchanged.
\Cref{tab:discourse_relations} offers an example where world knowledge must be taken into account for coreference. It is also worth noting that the choice of antecedent for the ``it'' in utterance (b) has no influence on the CT-based coherence scores.

\section{Other Related Work}
Some previous works have also explored the connection between CT and coreference. \citet{sidner1979towards} investigated the process of focussing
in the context of comprehension of anaphoric expressions in English discourse. \citet{brennan-etal-1987-centering} presented a formalization of centering and used it as the basis for an algorithm to track discourse context and bind pronouns, called the BFP algorithm. Then a critical evaluation of CT for pronoun interpretation was proposed by~\citet{kehler1997current}.
\citet{tetreault-2001-corpus} then compared pronoun resolution algorithms and introduced a centering algorithm (LeftRight Centering). After that, \citet{beaver2004optimization} reformulated the centering model of anaphora resolution and discourse coherence. Later, \citet{kong-etal-2009-employing} employed CT in pronoun resolution from a semantic perspective. 
An interesting review on the connection between centering and anaphora can be found in \citet{JOSHI2006223}.
The most recent related work would be \cite{kibble-power-2000-integrated}, which tries to incorporates CT into neural coreference resolution and improves its performance, especially on pronoun resolution in long documents, formal well-structured text, and clusters with scattered mentions.
While these works have attempted to use CT or some modified version of it for coreference, CT itself is not a theory of coreference resolution. The systematic analysis in this paper attempts to address the issue of the under-explored correlation of CT and coreference and provides a further theoretical basis for the use of CT in coreference. Moreover, we explore the relationship in the context of modern coreference systems and a much larger dataset.

\section{Conclusion}\label{conc}
By building a joint computational framework of CT and coreferece, we conclude that although CT itself is not a theory of coreference,
there exists a strong dependency between coreference and CT.
However, this dependence is not linear. When the coreference quality of a model is high enough, the usefulness of CT is limited.
We also confirm that neural coreference models, especially those adopting contextualized embeddings, contain much discourse-level coherence information. 
Thus, we conclude that CT can only provide minimal gains to modern coreference models.
That being said, the low mutual information between coherence and coreference suggests that coherence modeling should still be incorporated into neural models, given that it cannot be learned entirely from coreference annotations.
Finally, we explore which linguistic factors contribute to anaphora and may not be captured by CT; and find that recency bias, discourse relations and world knowledge may help explain some of the difference.

\bibliography{acl2020} 

\begin{thebibliography}{34}
\expandafter\ifx\csname natexlab\endcsname\relax\def\natexlab#1{#1}\fi

\bibitem[{Barzilay and Lapata(2008)}]{barzilay-lapata-2008-modeling}
Regina Barzilay and Mirella Lapata. 2008.
\newblock \href {https://doi.org/10.1162/coli.2008.34.1.1} {Modeling local
  coherence: An entity-based approach}.
\newblock \emph{Computational Linguistics}, 34(1):1--34.

\bibitem[{Beaver(2004)}]{beaver2004optimization}
David~I. Beaver. 2004.
\newblock \href
  {https://link.springer.com/article/10.1023/B:LING.0000010796.76522.7a} {The
  optimization of discourse anaphora}.
\newblock \emph{Linguistics and Philosophy}, 27(1):3--56.

\bibitem[{Brennan et~al.(1987)Brennan, Friedman, and
  Pollard}]{brennan-etal-1987-centering}
Susan~E. Brennan, Marilyn~W. Friedman, and Carl~J. Pollard. 1987.
\newblock \href {https://doi.org/10.3115/981175.981197} {A centering approach
  to pronouns}.
\newblock In \emph{25th Annual Meeting of the Association for Computational
  Linguistics}, pages 155--162, Stanford, California, USA. Association for
  Computational Linguistics.

\bibitem[{Chai and Strube(2022)}]{chai-strube-2022-incorporating}
Haixia Chai and Michael Strube. 2022.
\newblock \href {https://doi.org/10.18653/v1/2022.naacl-main.218}
  {Incorporating centering theory into neural coreference resolution}.
\newblock In \emph{Proceedings of the 2022 Conference of the North American
  Chapter of the Association for Computational Linguistics: Human Language
  Technologies}, pages 2996--3002, Seattle, United States. Association for
  Computational Linguistics.

\bibitem[{Chambers and Smyth(1998)}]{chambers1998structural}
Craig~G. Chambers and Ron Smyth. 1998.
\newblock \href
  {https://www.sciencedirect.com/science/article/abs/pii/S0749596X9892575X}
  {Structural parallelism and discourse coherence: A test of centering theory}.
\newblock \emph{Journal of Memory and Language}, 39(4):593--608.

\bibitem[{van Deemter and Kibble(2000)}]{deemter2000coreferring}
Kees van Deemter and Rodger Kibble. 2000.
\newblock \href {https://www.aclweb.org/anthology/J00-4005} {On coreferring:
  Coreference in {MUC} and related annotation schemes}.
\newblock \emph{Computational Linguistics}, 26(4):629--637.

\bibitem[{Elango(2005)}]{elango2005coreference}
Pradheep Elango. 2005.
\newblock \href
  {https://ccc.inaoep.mx/~villasen/index_archivos/cursoTATII/EntidadesNombradas/Elango-SurveyCoreferenceResolution.pdf}
  {Coreference resolution: A survey}.
\newblock \emph{University of Wisconsin, Madison, WI}.

\bibitem[{Gernsbacher(1990)}]{gernsbacher-1990-language}
Morton~Ann Gernsbacher. 1990.
\newblock \href
  {https://www.routledge.com/Language-Comprehension-As-Structure-Building/Gernsbacher/p/book/9780805806762}
  {\emph{Language comprehension as structure building}}.
\newblock Lawrence Erlbaum Associates.

\bibitem[{Gernsbacher et~al.(1989)Gernsbacher, Hargreaves, and
  Beeman}]{gernsbacher-1989-building}
Morton~Ann Gernsbacher, David~J Hargreaves, and Mark Beeman. 1989.
\newblock \href {https://www.ncbi.nlm.nih.gov/pmc/articles/PMC4260528}
  {Building and accessing clausal representations: The advantage of first
  mention versus the advantage of clause recency}.
\newblock \emph{Journal of memory and language}, 28(6):735--755.

\bibitem[{Gordon et~al.(1993)Gordon, Grosz, and Gilliom}]{gordon1993pronouns}
Peter~C. Gordon, Barbara~J. Grosz, and Laura~A. Gilliom. 1993.
\newblock \href
  {https://onlinelibrary.wiley.com/doi/abs/10.1207/s15516709cog1703_1}
  {Pronouns, names, and the centering of attention in discourse}.
\newblock \emph{Cognitive Science}, 17(3):311--347.

\bibitem[{Gordon and Hendrick(1998)}]{gordon1998representation}
Peter~C. Gordon and Randall Hendrick. 1998.
\newblock \href
  {https://www.sciencedirect.com/science/article/abs/pii/S0364021399800457}
  {The representation and processing of coreference in discourse}.
\newblock \emph{Cognitive Science}, 22(4):389--424.

\bibitem[{Grosz et~al.(1995)Grosz, Joshi, and
  Weinstein}]{grosz-etal-1995-centering}
Barbara~J. Grosz, Aravind~K. Joshi, and Scott Weinstein. 1995.
\newblock \href {https://www.aclweb.org/anthology/J95-2003} {{C}entering: A
  framework for modeling the local coherence of discourse}.
\newblock \emph{Computational Linguistics}, 21(2):203--225.

\bibitem[{Hobbs(1978)}]{hobbs1978resolving}
Jerry~R. Hobbs. 1978.
\newblock \href
  {https://www.sciencedirect.com/science/article/pii/0024384178900062}
  {Resolving pronoun references}.
\newblock \emph{Lingua}, 44(4):311--338.

\bibitem[{Hovy et~al.(2006)Hovy, Marcus, Palmer, Ramshaw, and
  Weischedel}]{weischedel2013ontonotes}
Eduard Hovy, Mitchell Marcus, Martha Palmer, Lance Ramshaw, and Ralph
  Weischedel. 2006.
\newblock \href {https://aclanthology.org/N06-2015} {{O}nto{N}otes: The 90{\%}
  solution}.
\newblock In \emph{Proceedings of the Human Language Technology Conference of
  the {NAACL}, Companion Volume: Short Papers}, pages 57--60, New York City,
  USA. Association for Computational Linguistics.

\bibitem[{Iida et~al.(2003)Iida, Inui, Takamura, and
  Matsumoto}]{iida2003incorporating}
Ryu Iida, Kentaro Inui, Hiroya Takamura, and Yuji Matsumoto. 2003.
\newblock \href {https://www.aclweb.org/anthology/W03-2604} {Incorporating
  contextual cues in trainable models for coreference resolution}.
\newblock In \emph{Proceedings of the 2003 {EACL} Workshop on The Computational
  Treatment of Anaphora}.

\bibitem[{Joshi et~al.(2006)Joshi, Prasad, and Miltsakaki}]{JOSHI2006223}
A.~K. Joshi, R.~Prasad, and E.~Miltsakaki. 2006.
\newblock \href
  {https://onesearch.library.rice.edu/discovery/fulldisplay?docid=cdi_elsevier_sciencedirect_doi_10_1016_B0_08_044854_2_04366_2\&context=PC\&vid=01RICE_INST:RICE&lang=en\&adaptor=Primo\%20Central\&tab=Everything\&query=null,,English,AND\&facet=citing,exact,cdi_FETCH-LOGICAL-14901-8cfec6daa36d841c44018663c65ddc8f845e50ab9216f6ed93a34b1668f7f63}
  {Anaphora resolution: Centering theory approach}.
\newblock In Keith Brown, editor, \emph{Encyclopedia of Language \&
  Linguistics}, 2 edition, pages 223 -- 230. Oxford.

\bibitem[{Joshi et~al.(2020)Joshi, Chen, Liu, Weld, Zettlemoyer, and
  Levy}]{joshi-etal-2020-spanbert}
Mandar Joshi, Danqi Chen, Yinhan Liu, Daniel~S. Weld, Luke Zettlemoyer, and
  Omer Levy. 2020.
\newblock \href {https://doi.org/10.1162/tacl_a_00300} {{S}pan{BERT}: Improving
  pre-training by representing and predicting spans}.
\newblock \emph{Transactions of the Association for Computational Linguistics},
  8:64--77.

\bibitem[{Karamanis et~al.(2004)Karamanis, Poesio, Mellish, and
  Oberlander}]{karamanis2004evaluating}
Nikiforos Karamanis, Massimo Poesio, Chris Mellish, and Jon Oberlander. 2004.
\newblock \href {https://doi.org/10.3115/1218955.1219005} {Evaluating
  centering-based metrics of coherence for text structuring using a reliably
  annotated corpus}.
\newblock In \emph{Proceedings of the 42nd Annual Meeting of the Association
  for Computational Linguistics ({ACL}-04)}, pages 391--398, Barcelona, Spain.

\bibitem[{Kehler(1997)}]{kehler1997current}
Andrew Kehler. 1997.
\newblock \href {https://aclanthology.org/J97-3006} {Current theories of
  centering for pronoun interpretation: a critical evaluation}.
\newblock \emph{Computational Linguistics}, 23(3):467--475.

\bibitem[{Kehler and Rohde(2013)}]{kehler2013probabilistic}
Andrew Kehler and Hannah Rohde. 2013.
\newblock \href
  {https://www.degruyter.com/document/doi/10.1515/tl-2013-0001/html} {A
  probabilistic reconciliation of coherence-driven and centering-driven
  theories of pronoun interpretation}.
\newblock \emph{Theoretical Linguistics}, 39(1-2):1--37.

\bibitem[{Kibble and Power(2000)}]{kibble-power-2000-integrated}
Rodger Kibble and Richard Power. 2000.
\newblock \href {https://doi.org/10.3115/1118253.1118265} {An integrated
  framework for text planning and pronominalisation}.
\newblock In \emph{{INLG}{'}2000 Proceedings of the First International
  Conference on Natural Language Generation}, pages 77--84, Mitzpe Ramon,
  Israel. Association for Computational Linguistics.

\bibitem[{Kong et~al.(2009)Kong, Zhou, and Zhu}]{kong-etal-2009-employing}
Fang Kong, Guodong Zhou, and Qiaoming Zhu. 2009.
\newblock \href {https://www.aclweb.org/anthology/D09-1103} {Employing the
  centering theory in pronoun resolution from the semantic perspective}.
\newblock In \emph{Proceedings of the 2009 Conference on Empirical Methods in
  Natural Language Processing}, pages 987--996, Singapore. Association for
  Computational Linguistics.

\bibitem[{Lee et~al.(2017)Lee, He, Lewis, and Zettlemoyer}]{lee-etal-2017-end}
Kenton Lee, Luheng He, Mike Lewis, and Luke Zettlemoyer. 2017.
\newblock \href {https://doi.org/10.18653/v1/D17-1018} {End-to-end neural
  coreference resolution}.
\newblock In \emph{Proceedings of the 2017 Conference on Empirical Methods in
  Natural Language Processing}, pages 188--197, Copenhagen, Denmark.
  Association for Computational Linguistics.

\bibitem[{Lee et~al.(2018)Lee, He, and Zettlemoyer}]{lee-etal-2018-higher}
Kenton Lee, Luheng He, and Luke Zettlemoyer. 2018.
\newblock \href {https://doi.org/10.18653/v1/N18-2108} {Higher-order
  coreference resolution with coarse-to-fine inference}.
\newblock In \emph{Proceedings of the 2018 Conference of the North {A}merican
  Chapter of the Association for Computational Linguistics: Human Language
  Technologies, Volume 2 (Short Papers)}, pages 687--692, New Orleans,
  Louisiana. Association for Computational Linguistics.

\bibitem[{Pennington et~al.(2014)Pennington, Socher, and
  Manning}]{pennington2014glove}
Jeffrey Pennington, Richard Socher, and Christopher~D. Manning. 2014.
\newblock \href {http://www.aclweb.org/anthology/D14-1162} {{GloVe}: Global
  vectors for word representation}.
\newblock In \emph{Empirical Methods in Natural Language Processing (EMNLP)},
  pages 1532--1543.

\bibitem[{Poesio et~al.(2004)Poesio, Stevenson, Di~Eugenio, and
  Hitzeman}]{poesio-etal-2004-centering}
Massimo Poesio, Rosemary Stevenson, Barbara Di~Eugenio, and Janet Hitzeman.
  2004.
\newblock \href {https://doi.org/10.1162/0891201041850911} {{C}entering: A
  parametric theory and its instantiations}.
\newblock \emph{Computational Linguistics}, 30(3):309--363.

\bibitem[{Pradhan et~al.(2013)Pradhan, Moschitti, Xue, Ng, Bj{\"o}rkelund,
  Uryupina, Zhang, and Zhong}]{pradhan2013towards}
Sameer Pradhan, Alessandro Moschitti, Nianwen Xue, Hwee~Tou Ng, Anders
  Bj{\"o}rkelund, Olga Uryupina, Yuchen Zhang, and Zhi Zhong. 2013.
\newblock \href {https://www.aclweb.org/anthology/W13-3516} {Towards robust
  linguistic analysis using {O}nto{N}otes}.
\newblock In \emph{Proceedings of the Seventeenth Conference on Computational
  Natural Language Learning}, pages 143--152, Sofia, Bulgaria. Association for
  Computational Linguistics.

\bibitem[{Sidner(1979)}]{sidner1979towards}
Candace~Lee Sidner. 1979.
\newblock \href {https://dspace.mit.edu/handle/1721.1/6880} {Towards a
  computational theory of definite anaphora comprehension in english
  discourse}.
\newblock Technical report, Massachusetts Institute of Technology.

\bibitem[{Strube and Hahn(1999)}]{strube-hahn-1999-functional}
Michael Strube and Udo Hahn. 1999.
\newblock \href {https://aclanthology.org/J99-3001} {Functional centering {--}
  grounding referential coherence on information structure}.
\newblock \emph{Computational Linguistics}, 25(3):309--344.

\bibitem[{Tetreault(2001)}]{tetreault-2001-corpus}
Joel~R. Tetreault. 2001.
\newblock \href {https://www.aclweb.org/anthology/J01-4003} {A corpus-based
  evaluation of centering and pronoun resolution}.
\newblock \emph{Computational Linguistics}, 27(4):507--520.

\bibitem[{Walker et~al.(1994)Walker, Iida, and
  Cote}]{walker-etal-1994-japanese}
Marilyn Walker, Masayo Iida, and Sharon Cote. 1994.
\newblock \href {https://aclanthology.org/J94-2003} {{J}apanese discourse and
  the process of centering}.
\newblock \emph{Computational Linguistics}, 20(2):193--231.

\bibitem[{Walker(1998)}]{walker1998centering}
Marilyn~A. Walker. 1998.
\newblock \href {https://arxiv.org/abs/cmp-lg/9708005} {Centering, anaphora
  resolution, and discourse structure}.
\newblock \emph{Centering Theory in Discourse}, pages 401--435.

\bibitem[{Wu et~al.(2020)Wu, Wang, Yuan, Wu, and Li}]{wu-etal-2020-corefqa}
Wei Wu, Fei Wang, Arianna Yuan, Fei Wu, and Jiwei Li. 2020.
\newblock \href {https://doi.org/10.18653/v1/2020.acl-main.622} {{C}oref{QA}:
  Coreference resolution as query-based span prediction}.
\newblock In \emph{Proceedings of the 58th Annual Meeting of the Association
  for Computational Linguistics}, pages 6953--6963, Online. Association for
  Computational Linguistics.

\bibitem[{Xia et~al.(2020)Xia, Sedoc, and
  Van~Durme}]{xia-etal-2020-incremental}
Patrick Xia, Jo{\~a}o Sedoc, and Benjamin Van~Durme. 2020.
\newblock \href {https://doi.org/10.18653/v1/2020.emnlp-main.695} {Incremental
  neural coreference resolution in constant memory}.
\newblock In \emph{Proceedings of the 2020 Conference on Empirical Methods in
  Natural Language Processing (EMNLP)}, pages 8617--8624, Online. Association
  for Computational Linguistics.

\end{thebibliography}
\bibliographystyle{acl_natbib.bst}

\end{document}



\begin{appendices}
\section{Realization in Centering Theory.}
The definition of \textit{realize} depends on the semantic theory one adopts.
\paragraph{Direct realization} $U$ directly realizes $c$ where $U$ is an utterance of some phrase for which $c$ is the semantic interpretation. NP directly realizes $c$ can hold in cases where NP is a definite description and $c$ is its denotation, its value-free interpretation, e.g. “the current US president” directly realizes “Donald Trump”.


\paragraph{Indirect realization}  $U$ indirectly realizes $c$ when one of the noun phrases
in $U$ is an associative reference to that discourse entity \cite{hawkins2015definiteness}. The generalization from direct realization to indirect realization allows capturing certain regularities in the use of definite descriptions and pronouns.

\section{Reformulating Coreference Models in CT Framework as $\mathbb{P}(\mathcal{E} \mid \mathcal{D})$.}
In this appendix, we reformulate the state-of-the-art neural coreference resolution models in the proposed CT framework. 
\subsection{\textit{e2e} Model}
The most popular end-to-end coreference framework, \textit{e2e}~\cite{lee-etal-2017-end}, can be seen as a latent variable model.
They obtain a set of candidate mention spans $\mathcal{M}_i$ for each discourse by considering spans until a certain length, those that do not cross sentence boundaries, etc.

They then implement the mention-entity mapping function $f$ by a chain-linking algorithm. For each discourse $\mathcal{D}_i (i=1 \dots N)$, we define a sequence of latent variables $\bm{A}_i= \langle \bm{a}_{i,1}, \ldots, \bm{a}_{i,p} \rangle$, where $\bm{a}_{i,j}$ takes values in $\mathcal{A}(i,j) = \{ 0, 1, \dots, j-1\} $. Here, $0$ is reserved for $\epsilon$ i.e. ``new cluster'' or ``not a entity'' or ``not a mention''.

They aim to learn a conditional probability distribution
$\mathbb{P}(\bm{A}_1, \ldots , \bm{A}_N| \mathcal{D})$ whose most likely configuration produces the correct clustering for each discourse.
They use a product of multinomials for each span:
\begin{equation}
    \mathbb{P}(\bm{A}_1, \ldots , \bm{A}_N| \mathcal{D}) = \prod_{i=1}^{N} \prod_{j=1}^{|\mathcal{M}_i|} \mathbb{P}(\bm{a}_{i,j}| \mathcal{D}_i).
    \notag
\end{equation}
They further model this probability in the form of a softmax:
\begin{equation}
\begin{aligned}
    &\mathbb{P}(\bm{a}_{i,j}| \mathcal{D}_i) =
    &\prod_{j=1}^{|\mathcal{M}_i|} \frac{\exp(s(m_{i,j}, a_{i,j}))}{\sum_{a'\in \mathcal{A}(i,j)} \exp(s(m_{i,j}, a'))},
    \notag
\end{aligned}
\end{equation}
where $s(m_{i,j}, a_{i,j})$ is a pairwise score for a coreference link between mention span $m_{i,j}$ and $a_{i,j}$ in discourse $\mathcal{D}$ as
\begin{equation}
    \left\{
    \begin{array}{ll}
       s_m(m_{i,j}) + s_m(a_{i,j}) + s_a(m_{i,j}, a_{i,j}) &  a_{i,j}\neq \epsilon\\
       0  &  a_{i,j}= \epsilon
    \end{array} \right.
    \notag
\end{equation}
Here $s_m(m_{i,j})$ is a unary score for span $m_{i,j}$ being a mention, and $s_a(m_{i,j}, a_{i,j})$ is pairwise score for span $a_{i,j}$ being an antecedent of span $m_{i,j}$.
The scoring functions are computed via feed-forward neural networks as:
\begin{equation}
\begin{aligned}
    s_m(m_{i,j}) = & \bm{w}_m \cdot f_m(\bm{g}(m_{i,j})), \\
    s_a(m_{i,j}, a_{i,j}) = & \bm{w}_a \cdot f_a(\bm{g}(m_{i,j}), \bm{g}(a_{i,j}),\\
    &\bm{g}(m_{i,j}) \circ  \bm{g}(a_{i,j}),  \phi(i,j))
\end{aligned}
\notag
\end{equation}
The antecedent scoring function $s_a(m_{i,j}, a_{i,j})$ includes explicit element-wise similarity of each span $ \bm{g}(m_{i,j}) \circ  \bm{g}(a_{i,j})$ and a feature vector $\phi(i,j)$ encoding speaker and genre information from the metadata and the distance between the two spans.
\paragraph{Learning}: They learn their model by maximizing the marginal log-likelihood of all correct antecedents implied by the gold clustering as
\begin{equation}
    \log \prod_{i=1}^{N} \prod_{j=1}^{|M_i|} \sum_{\hat{a}\in \mathcal{A}(i,j)\cap GOLD(i,j)} \mathbb{P}(\hat{a} | \mathcal{D}).
    \notag
\end{equation}

\subsection{Higher Order Model}
In the \textit{e2e} model, only \textit{pairs} of entity mentions are scored by the model. It makes independent decisions about coreference links instead of adding mentions to clusters globally. \textit{c2f}~\cite{lee-etal-2018-higher} improves the first order model to higher orders. They use the antecedent distribution from a span-ranking architecture as an attention mechanism to iteratively refine
span representations. With the definitions of $\mathbb{P}(\bm{a}_{i,j}| \mathcal{D}_i)$, $s(m_{i,j}, a_{i,j})$, $s_m(m_{i,j})$ and $s_a(m_{i,j}, a_{i,j})$ being the same, the span representations $\bm{g}(m_{i,j})$ and $\bm{g}(a_{i,j})$ are computed via iterations:
\begin{equation}
\begin{aligned}
\bm{h}_{i,j}  = &  \sum\limits_{a' \in \mathcal{A}(i,j)} \mathbb{P}(a'| \mathcal{D}_i)  \cdot   \bm{g}(a')^{t} \\
\bm{g}(m_{i,j})^{t+1}  = & f_h(\bm{g}(m_{i,j}))^t \circ \bm{g}(m_{i,j})^t \\
& + (1- f_h(\bm{g}(m_{i,j})^t) ) \circ \bm{h}_{i,j} \\
\bm{g}(a_{i,j})^{t+1}  = & f_h(\bm{g}(a_{i,j}))^t \circ \bm{g}(a_{i,j})^t \\
& + (1- f_h(\bm{g}(a_{i,j})^t) ) \circ \bm{h}_{i,j}
\end{aligned}
\notag
\end{equation}
where $\bm{g}(m_{i,j})^{t}$ and $\bm{g}(a_{i,j})^{t}$ are the span representations of iteration $t$. The initialization of $\bm{g}(m_{i,j})^1$ and $\bm{g}(a_{i,j})^1$ is done by the first order model.

They also introduce a coarse-to-fine approach to alleviate the computational cost.

\subsection{Incremental Coreference Model}
\cite{xia-etal-2020-incremental} implements the mention-entity mapping function $f$ by a incremental algorithm as introduced in Algorithm \ref{a1}.
\begin{algorithm} 
\SetAlgoLined
 Create an empty Entity Set $\mathcal{E}_i$\;
 \For{$U \in \mathcal{D}_i$}{
  $M \leftarrow$ SPANS$(U)$\;
  \For{$m \in M$}{
  scores $\leftarrow$ PAIRSCORE$(m,\mathcal{E}_i)$\;
  top\_score $\leftarrow$ max(scores)\;
  top\_e $\leftarrow$ argmax(scores)\;
\If{\rm top\_score $>0$}{
   Update(top\_e, $m$)\;
   }{
   AddNewEntity($\mathcal{E}_i, m$)\;
  }
  }
  EVICT($\mathcal{E}_i$)
 }
 \caption{FindCluster($\mathcal{D}_i$)}
 \label{a1}
\end{algorithm}

They attach each $e \in \mathcal{E}_i$ with its own representation. For each span, a scorer scores the span representation against all the cluster representations. This is used to determine to which (if any) of the pre-existing $e \in \mathcal{E}_i$ the current span should be mapped to. Upon inclusion of the span in the cluster, the $e$’s representation is subsequently updated via a (learned) function. Periodically, the model evicts less salient entities, writing them to disk (based on cluster size and distance from the end of the segment). Under this algorithm, each clustering decision is permanent.

\section{Coreference Model Parameters Settings.}
\subsection{Parameters Settings}
For \textit{c2f-SpanBERT} models, we follow the implementation in \cite{joshi-etal-2019-bert}. We use the cased English variants in all our experiments. We set the span to word ratio (\texttt{spans\_per\_word}) to 0.5, 0.4 or 0.2 for different models. Documents are split into segments of \texttt{max\_segment\_len} of 512 for SpanBERT-large and 384 for SpanBERT-base. The dimension of all feed-forward networks is 1500 for SpanBERT-large and 1000 for SpanBERT-base. The most of models adopt \texttt{inference\_order} of 1, while for the best performing ones we implemented we set it to 2 with \texttt{coarse\_to\_fine} being true. 

For \textit{c2f-one-hot} models and \textit{c2f-GloVe} models, we follow the implementation in \cite{lee-etal-2018-higher}. \textit{c2f-GloVe} models take GloVe embeddings as input while \textit{c2f-one-hot} models use one-hot word vectors as input and does not levarage any pretrained embeddings. We set the LSTM context layer to be 1 layer. We set the span to word ratio (\texttt{spans\_per\_word}) to 0.4. The dimension of all feed-forward networks is 1500. They all adopt \texttt{inference\_order} of 1.

For all the models mentioned above, we set the upper bound number of spans to \texttt{max\_antecedents} of 50. We also filtered out spans that are longer than 30 tokens. The dimension of the feature fed into feed-forward networks is 20. 

For \textit{corefQA}~\cite{wu-etal-2020-corefqa} and \textit{Incremental}~\cite{xia-etal-2020-incremental}, we follow the parameter settings specified in their papers.

\subsection{Training Details}
We trained models with Allennlp~\cite{Gardner2017AllenNLP} on the OntoNotes dataset for 40 epochs, using a dropout rate of 0.3. We adopted early stopping with patience being 10. We used a batch size of 1 document for SpanBERT-large and 4 for SpanBERT-base. SpanBERT-large models are trained on 4 Nvidia Tesla V100 GPUs while other models are trained on 1 Nvidia Tesla V100 GPU.

\subsection{Two models in Section 7.2.}
In Section 7.2, we compare different center-based metrics based on the CF-candidates provided by two coreference models, a strong one (well-trained \textit{c2f-SpanBERT-large}, with a coref F1 of 78.87\%) and a weak one (underfitted \textit{e2e-SpanBERT-base}, with a coref F1 of 17.51\%), under four instantiations of CT. These two models follow the parameters settings for SpanBERT-large and SpanBERT-base in C.1. The \texttt{inference\_order} of \textit{c2f-SpanBERT-large} is set to 2; the \texttt{inference\_order} of \textit{e2e-SpanBERT-base} is set to 1.  \textit{c2f-SpanBERT-large} is trained for 40 epochs till convergence, while \textit{e2e-SpanBERT-base} is trained only for 2 epochs.

\subsection{Checkpoints in Section 7.3.}
We plot all epoches of all types of embeddings (SpanBERT, GloVe, One-hot) on all different sizes of datasets ($10\%, 20\%, \cdots, 100\%$ of the original training set):
\begin{itemize}
    \item \textbf{c2f-base}: 40 epoches * 3 types of embedding * 10 datasets * 5 runs = 6000
    \item \textbf{c2f-large-SpanBERT}: 40 epoches * 10 datasets * 5 runs = 2000
    \item \textbf{pretrained-c2f-large-SpanBERT}: 40 epoches * 5 runs = 200
     \item \textbf{corefQA}: 40 epoches * 5 runs = 200
      \item \textbf{Incremental}: 40 epoches * 5 runs = 200
\end{itemize}

In total, there are 8,600 checkpoints. The best epoches of \textbf{c2f-large-SpanBERT}, \textbf{pretrained-c2f-large-SpanBERT}, \textbf{corefQA} and \textbf{Incremental} trained on the entire training set (100\%) are shown in Figure 1, referred as the state-of-the-art models in recent years.

\section{Corpus-based Analysis of Centering Theory.}
Table \ref{tab:gold_cluster_grl_table_1} and Table \ref{tab:best_cluster_srl_table_1} present detailed statistics of the CT metrics on OntoNotes 5.0 test set. \textit{\#valid\_u} is the number of valid utterances (utterences with non-empty \texttt{CF-list candidates}) per document. \#M and \%M are the number of utterances which do not violate the metric M and the percentage of those utterances out of all valid utterances. \textit{Coh} is short for \textit{Coherence}, \textit{Sal} is short for \textit{Salience}, \textit{Cont.} is short for \textit{Continue}, \textit{Ret.} is short for \textit{Retain}, and $\overline{nocb}$ means the utterances with CB.

\begin{table*}[!thpb]
    \centering \small
    \begin{tabular}{c|c|c|c|c|c|c|c|c|c|c}
    \toprule[2pt]
      &  \#valid\_u&  \#nocb&  \#Cheap& \#Coh. & \#Sal. & \#Cont. & \#Ret.   & \#SShift & \#RShift & \#KP \\
    \midrule[1pt]
mean & 21.47 & 7.63 & 9.87 & 10.45 & 9.64 & 7.39 & 3.07 & 2.26 & 7.77 & 10.95\\ 
std & 18.22 & 6.40 & 10.16 & 10.12 & 9.58 & 7.51 & 3.36 & 2.75 & 7.13 & 10.74\\ 
min & 0.00 & 0.00 & 0.00 & 0.00 & 0.00 & 0.00 & 0.00 & 0.00 & 0.00 & 0.00\\ 
25\% & 7.00 & 2.00 & 2.00 & 3.00 & 3.00 & 2.00 & 1.00 & 0.00 & 2.00 & 3.00\\ 
50\% & 15.00 & 6.00 & 6.00 & 7.00 & 6.00 & 5.00 & 2.00 & 1.00 & 6.00 & 7.00\\ 
75\% & 31.25 & 11.00 & 14.00 & 15.00 & 14.00 & 11.00 & 5.00 & 3.00 & 12.00 & 15.25\\ 
max & 92.00 & 31.00 & 63.00 & 57.00 & 54.00 & 39.00 & 19.00 & 16.00 & 32.00 & 63.50\\ 
\midrule[1pt]
 & \%valid\_u & \%\overline{nocb} & \%Cheap & \%Coh. & \%Sal. & \%Cont. & \%Ret. & \%SShift & \%RShift & \%KP \\ 
\midrule[1pt]
mean & 80.02 & 60.14 & 42.78 & 46.51 & 42.85 & 39.17 & 14.15 & 9.99 & 36.12 & 48.07\\ 
std & 19.26 & 17.48 & 17.57 & 15.25 & 16.05 & 23.68 & 12.46 & 10.22 & 20.02 & 14.80\\ 
min & 0.00 & 0.00 & 0.00 & 0.00 & 0.00 & 0.00 & 0.00 & 0.00 & 0.00 & 0.00\\ 
25\% & 70.00 & 50.00 & 32.71 & 37.06 & 33.33 & 24.76 & 4.17 & 0.00 & 25.00 & 39.48\\ 
50\% & 84.62 & 61.45 & 44.44 & 47.11 & 44.44 & 35.56 & 12.50 & 9.09 & 36.36 & 50.00\\ 
75\% & 95.06 & 73.08 & 53.67 & 54.96 & 52.17 & 50.00 & 20.00 & 14.85 & 50.00 & 57.54\\ 
max & 100.00 & 93.48 & 87.50 & 87.50 & 87.50 & 100.00 & 100.00 & 50.00 & 100.00 & 87.50\\ 
\bottomrule[2pt]
\end{tabular}
\caption{The detailed statistics of CT-based metrics on OntoNotes, with CF candidates being ``cluster only'' and CF ranking being grammatical-role-based.}
\label{tab:gold_cluster_grl_table_1}
\end{table*}

\begin{table*}[!thpb]
    \centering  \small
    \begin{tabular}{c|c|c|c|c|c|c|c|c|c|c}
    \midrule[1pt]
 & \#valid\_u & \#nocb & \#Cheap & \#Coh. & \#Sal. & \#Cont. & \#Ret. & \#SShift & \#RShift & \#KP \\ 
    \toprule[2pt]
mean & 21.47 & 7.63 & 9.60 & 10.22 & 8.79 & 6.68 & 3.54 & 2.10 & 8.15 & 10.61\\ 
std & 18.22 & 6.40 & 9.86 & 9.88 & 8.67 & 6.83 & 3.82 & 2.52 & 7.46 & 10.36\\ 
min & 0.00 & 0.00 & 0.00 & 0.00 & 0.00 & 0.00 & 0.00 & 0.00 & 0.00 & 0.00\\ 
25\% & 7.00 & 2.00 & 2.75 & 3.00 & 2.00 & 2.00 & 1.00 & 0.00 & 2.00 & 3.00\\ 
50\% & 15.00 & 6.00 & 6.00 & 7.00 & 6.00 & 4.00 & 2.00 & 1.00 & 6.00 & 6.75\\ 
75\% & 31.25 & 11.00 & 14.00 & 15.00 & 13.00 & 10.00 & 5.00 & 3.00 & 13.00 & 15.56\\ 
max & 92.00 & 31.00 & 60.00 & 60.00 & 52.00 & 42.00 & 18.00 & 12.00 & 32.00 & 63.00\\ 
\midrule[1pt]
 & \%valid\_u & \%\overline{nocb} & \%Cheap & \%Coh. & \%Sal. & \%Cont. & \%Ret. & \%SShift & \%RShift & \%KP \\ 
\midrule[1pt]
mean & 80.02 & 60.14 & 41.62 & 45.79 & 39.67 & 36.50 & 16.10 & 9.23 & 37.59 & 46.81\\ 
std & 19.26 & 17.48 & 17.03 & 14.75 & 15.64 & 23.93 & 13.34 & 9.44 & 19.86 & 14.10\\ 
min & 0.00 & 0.00 & 0.00 & 0.00 & 0.00 & 0.00 & 0.00 & 0.00 & -0.00 & 0.00\\ 
25\% & 70.00 & 50.00 & 30.19 & 36.09 & 29.45 & 22.22 & 7.55 & 0.00 & 25.82 & 38.65\\ 
50\% & 84.62 & 61.45 & 42.86 & 46.15 & 40.00 & 33.33 & 15.38 & 8.33 & 38.46 & 47.79\\ 
75\% & 95.06 & 73.08 & 52.50 & 54.17 & 50.00 & 46.15 & 23.08 & 14.29 & 50.00 & 56.51\\ 
max & 100.00 & 93.48 & 80.00 & 85.71 & 80.00 & 100.00 & 100.00 & 50.00 & 100.00 & 80.00\\ 
    \bottomrule[2pt]
    \end{tabular}
\caption{The detailed statistics of CT-based metrics on OntoNotes, with CF candidates being ``cluster only'' and CF ranking being semantic-role-based.}
\label{tab:best_cluster_srl_table_1}
\end{table*}

\clearpage
\section{Coreference Performance of models}
Table \ref{tab:Coref} presents the complete results of coreference evaluation on the OntoNotes 5.0 test set. The metrics being used are MUC~\cite{vilain-etal-1995-model}, $B^3$~\cite{bagga1998algorithms} and $\mathrm{CEAF}_{\phi_4}$~\cite{luo-2005-coreference}. We include performance of systems proposed in the past 5 years for reference. We reevaluated the top performing models (e2e, c2e, Incremental and CorefQA), and report both the results directly from their papers and those of our implementations. Here we observed small differences in performance between our implementations and those reported in original papers, due to slightly different implementation details and parameter settings.

\begin{table}[!thpb]
    \centering \small
    \begin{tabular}{c|ccc|ccc|ccc|c}
    \toprule[2pt]
              &  \multicolumn{3}{c}{MUC} &  \multicolumn{3}{c}{$B^3$} & \multicolumn{3}{c}{$CEAF_{\phi_4}$} &  \\
        Model &  P & R & F1   &  P & R & F1   &  P & R & F1   &  Avg. F1  \\
    \midrule[1pt]
Martschat and Strube (2015)~\cite{martschat-strube-2015-latent} & 76.7 & 68.1 & 72.2 & 66.1 & 54.2 & 59.6 & 59.5 & 52.3 & 55.7 & 62.5\\
Clark and Manning (2015)~\cite{clark-manning-2015-entity} & 76.1 & 69.4 & 72.6 & 65.6 & 56.0 & 60.4 & 59.4 & 53.0 & 56.0 & 63.0\\
Wiseman et al. (2015)~\cite{wiseman-etal-2015-learning} & 76.2 & 69.3 & 72.6 & 66.2 & 55.8 & 60.5 & 59.4 & 54.9 & 57.1 & 63.4\\
Wiseman et al. (2016)~\cite{wiseman-etal-2016-learning} & 77.5 & 69.8 & 73.4 & 66.8 & 57.0 & 61.5 & 62.1 & 53.9 & 57.7 & 64.2\\
Clark and Manning (2016a)~\cite{clark-manning-2016-deep} & 79.9 & 69.3 & 74.2 & 71.0 & 56.5 & 63.0 & 63.8 & 54.3 & 58.7 & 65.3\\
Clark and Manning (2016b)~\cite{clark-manning-2016-improving} & 79.2 & 70.4 & 74.6 & 69.9 & 58.0 & 63.4 & 63.5 & 55.5 & 59.2 & 65.7\\
e2e~\cite{lee-etal-2017-end} & 78.4 & 73.4 & 75.8 & 68.6 & 61.8 & 65.0 & 62.7 & 59.0 & 60.8 & 67.2\\
c2f~\cite{lee-etal-2018-higher}& 81.4 & 79.5 & 80.4 & 72.2 & 69.5 & 70.8 & 68.2 & 67.1 & 67.6 & 73.0\\
    Fei et al. (2019)~\cite{fei-etal-2019-end} & 85.4 & 77.9 & 81.4 & 77.9 & 66.4 & 71.7 & 70.6 & 66.3 & 68.4 & 73.8\\
   EE~\cite{kantor-globerson-2019-coreference} & 82.6 & 84.1 & 83.4 & 73.3 & 76.2 & 74.7 & 72.4 & 71.1 & 71.8 & 76.6\\
\midrule[1pt]
e2e-BERT-base~\cite{joshi-etal-2019-bert} & 80.2 & 82.4 & 81.3 & 69.6 & 73.8 & 71.6 & 69.0 & 68.6 & 68.8 & 73.9\\
 c2f-BERT-large~\cite{joshi-etal-2019-bert} & 84.7 & 82.4 & 83.5 & 76.5 & 74.0 & 75.3 & 74.1 & 69.8 & 71.9 & 76.9\\
 Incremental-large~\cite{xia-etal-2020-incremental} & 85.7 & 84.8 & 85.3 & 78.1 & 77.5 & 77.8 & 76.3 & 74.1 & 75.2 & 79.4\\
 c2f-SpanBERT-large\cite{joshi-etal-2020-spanbert} & 85.8 & 84.8 & 85.3 & 78.3 & 77.9 & 78.1 & 76.4 & 74.2 & 75.3 & 79.6\\
 CorefQA-base~\cite{wu-etal-2020-corefqa} & 85.2 & 87.4 & 86.3 & 78.7 & 76.5 & 77.6 & 76.0 & 75.6 & 75.8 & 79.9 \\
 CorefQA-large~\cite{wu-etal-2020-corefqa} & 88.6 & 87.4 & 88.0 & 82.4 & 82.0 & 82.2 & 79.9 & 78.3 & 79.1 & 83.1 \\
 \midrule[1pt]
e2e-BERT-base (our) & 79.9 & 82.0 & 80.9 & 70.3 & 72.1 & 71.2 & 69.2 & 67.5 & 68.3 & 73.5\\
c2f-BERT-base (our) & 81.0 & 81.6 & 81.3 & 70.2 & 73.3 & 71.7 & 69.4 & 69.0 & 69.2 & 74.1\\
e2e-SpanBERT-base (our) & 85.6 & 82.5 & 84.0 & 76.8 & 74.9 & 75.8 & 75.9 & 70.2 & 72.9 & 77.6\\
c2f-SpanBERT-base (our) & 86.8 & 83.2 & 85.0 & 77.2 & 75.3 & 76.2 & 76.5 & 71.1 & 73.7 & 78.3\\
Incremental-base (our) & 85.2 & 82.9 & 84.0 & 77.0 & 75.1 & 76.0 & 75.4 & 70.5 & 72.9 & 77.6\\
CorefQA-base (our) & 85.0 & 86.2 & 85.6 & 78.9 & 75.6 & 77.2 & 76.2 & 76.8 & 76.5 & 79.8 \\
 \midrule[1pt]
e2e-BERT-large (our) & 83.9 & 84.3 & 84.1 & 75.5 & 74.2 & 74.8 & 73.0 & 69.0 & 70.9 & 76.6\\
c2f-BERT-large (our) & 84.7 & 82.4 & 83.5 & 76.5 & 74.0 & 75.3 & 74.1 & 69.8 & 71.9 & 76.9\\
e2e-SpanBERT-large (our) & 88.0 & 82.8 & 85.3 & 78.2 & 75.4 & 76.8 & 77.0 & 70.8 & 73.8 & 78.6\\
c2f-SpanBERT-large (our) & 84.9 & 85.2 & 85.0 & 77.3 & 77.0 & 77.1 & 78.0 & 76.2 & 77.1 & 79.8\\
Incremental-large (our) & 85.2 & 84.2 & 84.7 & 77.9 & 78.1 & 78.0 & 78.0 & 72.4 & 75.1 & 79.3\\
CorefQA-large (our) & 89.7 & 88.5 & 89.1 & 80.2 & 81.9 & 81.0 & 77.5 & 80.0 & 78.7 & 83.0\\
    \bottomrule[2pt]
    \end{tabular}
    \caption{Complete results of different coreference models on the OntoNotes 5.0 test set with three coreference resolution metrics: MUC, $B^3$, $CEAF_{\phi_4}$ and the average F1 of them. $P$, $R$ and $F1$ in the first row represent precision, recall and F1 score respectively. The bottom part is the results of our implementations. \textit{base} and \textit{large} represent the embedding encoder used.}
    \label{tab:Coref}
\end{table}

\clearpage

\section{Additional Analysis Results}
\begin{figure*}[thpb]
\includegraphics[scale = 0.28]{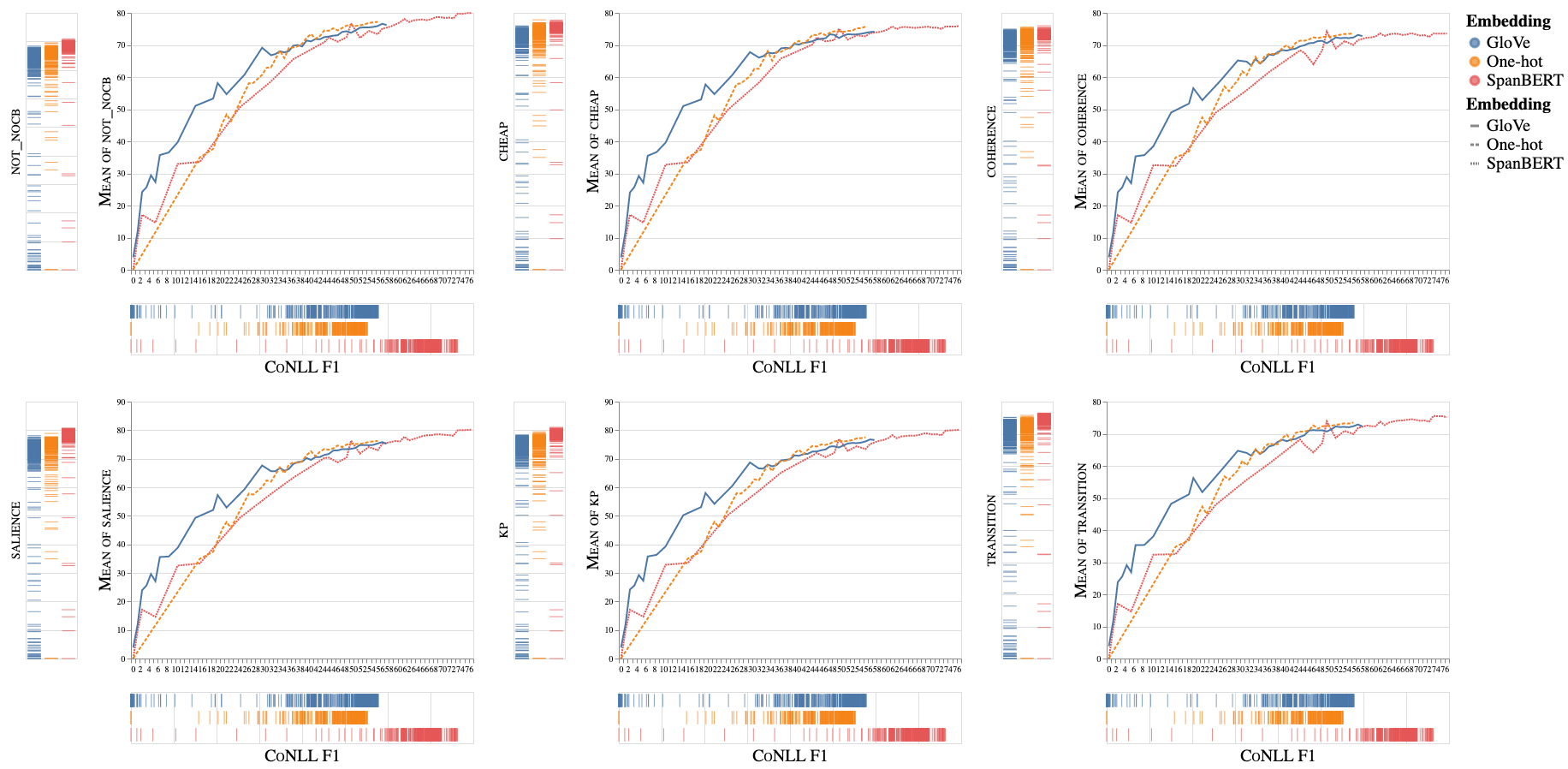}
\caption{The CT scores as a function of coreference F1 score for \textit{c2f-SpanBERT}, \textit{c2f-GloVe} and \textit{c2f-one-hot}.}
\label{fig:Glove-Spanbert-onehot-all-all}
\end{figure*}

\begin{figure*}[thpb]
\includegraphics[scale = 0.32]{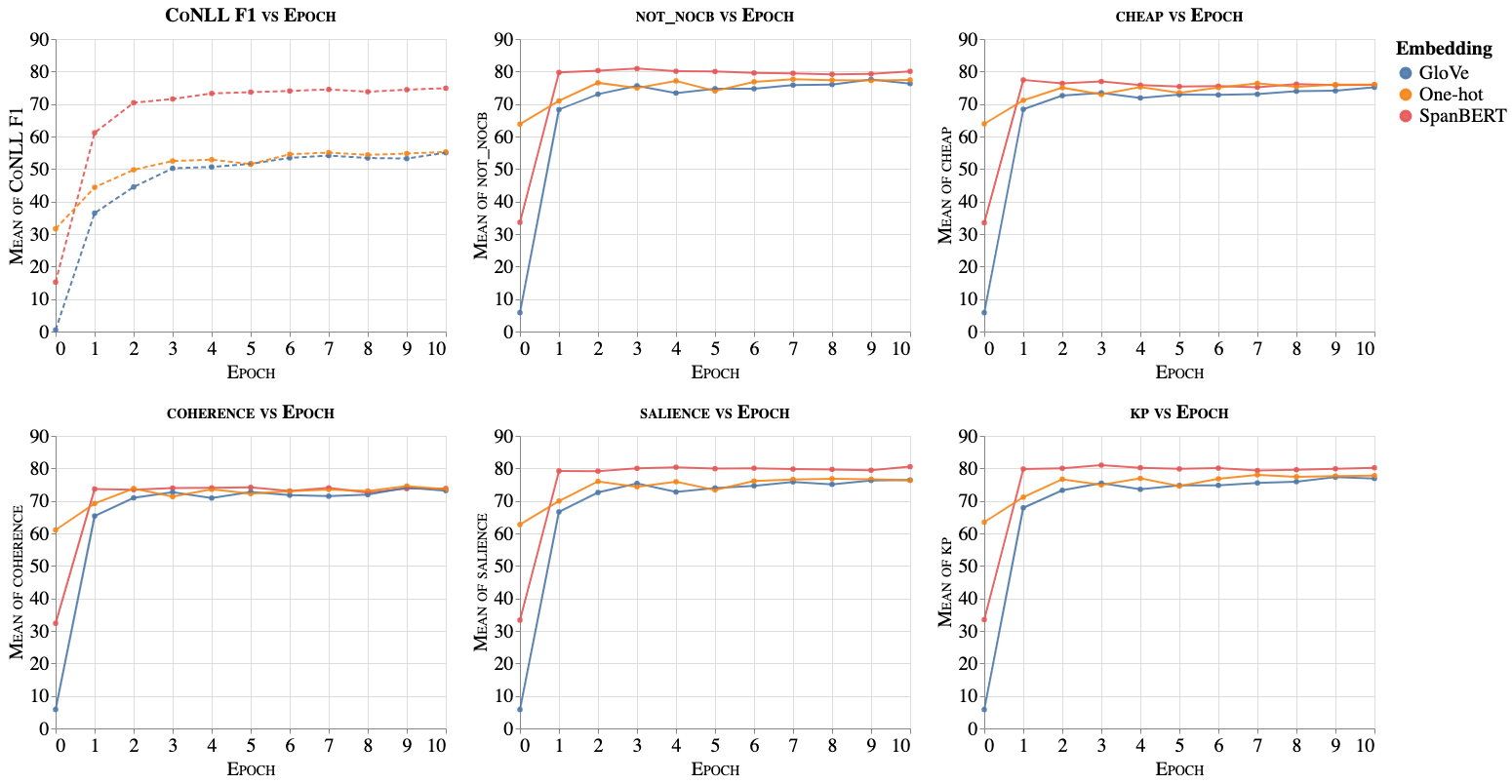}
\caption{The coref F1 score and the other 5 CT scores as a function of training epoch for \textit{c2f-SpanBERT}, \textit{c2f-GloVe} and \textit{c2f-one-hot}. Models are trained on 100\% OntoNotes training dataset.}
\label{fig:Glove-Spanbert-onehot-epoch-all}
\end{figure*}

\begin{figure*}[thpb]
\includegraphics[scale = 0.32]{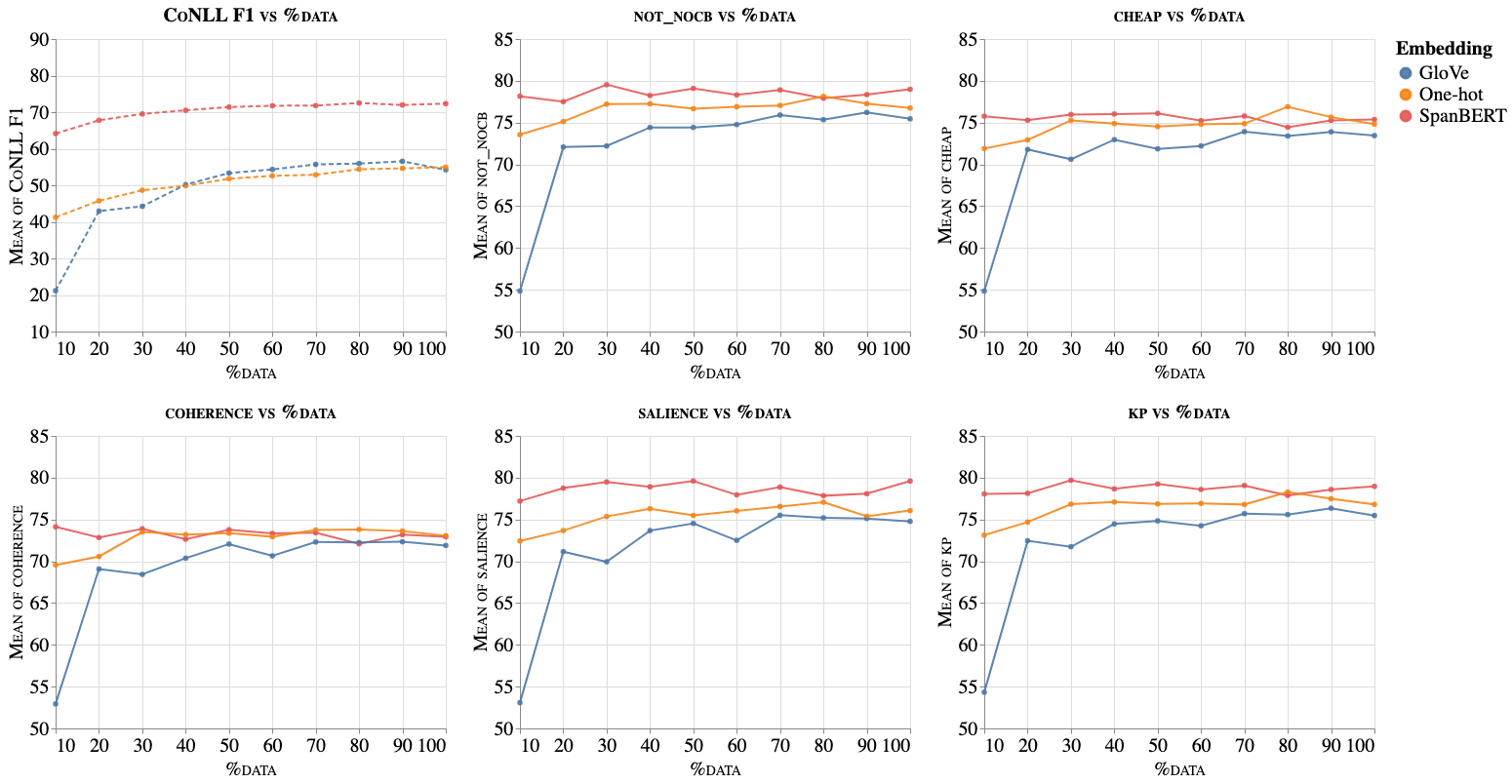}
\caption{The CT scores as a function of coreference F1 score for \textit{c2f-SpanBERT}, \textit{c2f-GloVe} and \textit{c2f-one-hot}.}
\label{fig:Glove-Spanbert-onehot-data-all}
\end{figure*}

\section{Analysis: Properties of Discourse.}
In this section, we exam whether different properties of discourse influence this relationship between coreference and Centering Theory.

\subsection{Document Length and \#Token per sentence}
\begin{figure*}[thpb]
\includegraphics[scale = 0.25]{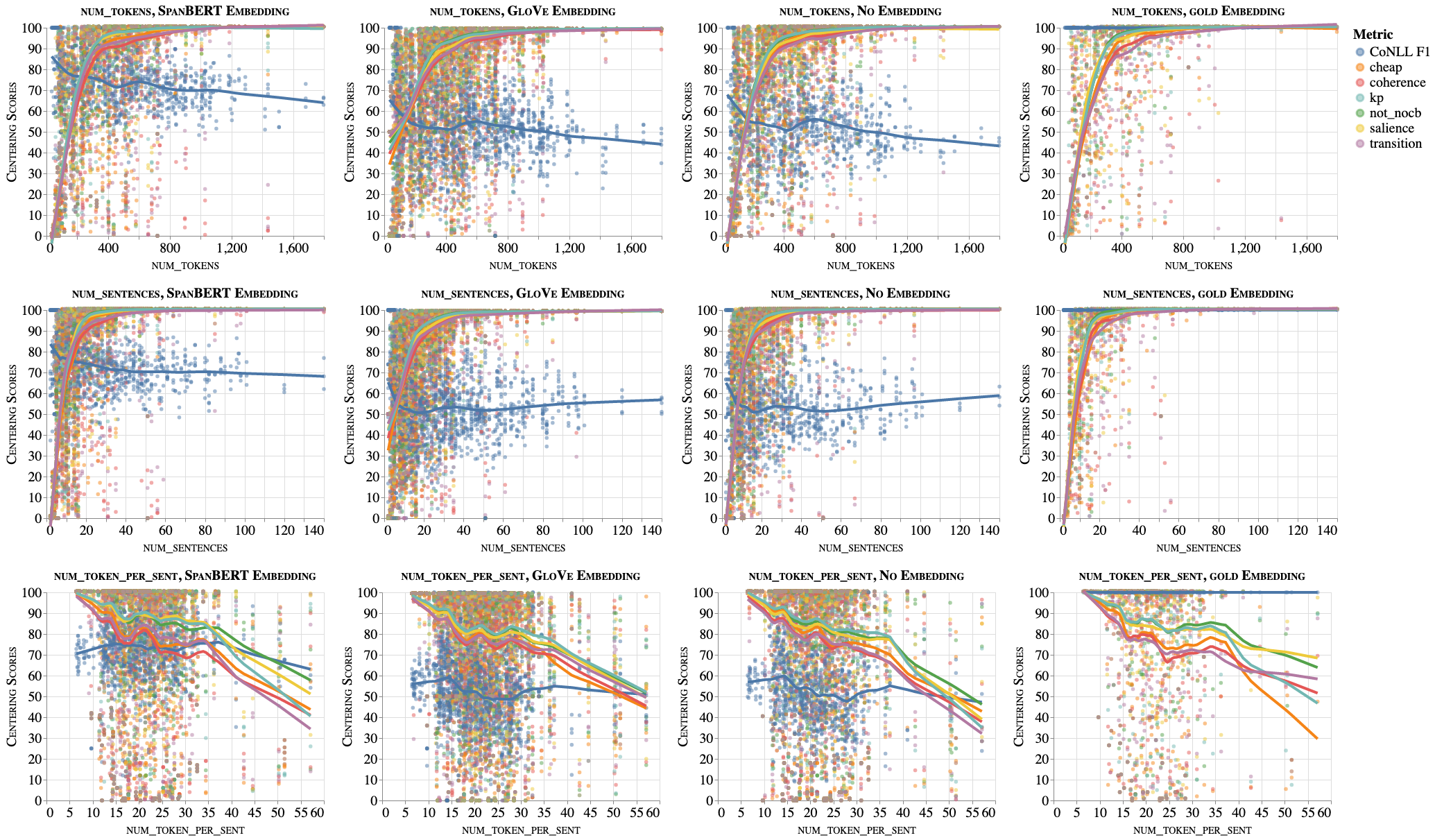}
\caption{The CT scores and the coreference F1 score as a function of discourse properties (i.g. \#sentences, \#tokens, average \#tokens per sentence) for \textit{c2f-SpanBERT}, \textit{c2f-GloVe} and \textit{c2f-one-hot} and Ontonotes ground truth annotations.}
\label{fig:Glove-Spanbert-onehot-document_length}
\end{figure*}

As we can see from Figure \ref{fig:Glove-Spanbert-onehot-document_length}, we observe that 1) the CT scores (80\%) is an average of all the documents, where longer documents tend to have higher CT scores and shorter documents have lower scores; 2) this is not the case for Coreference F1 score, where the predicted clusterings of shorter documents are actually better in the eye of Coreference resolution F1.
These figures show that for both the distribution of Y condictioned on the predicted $X$ and the real distribution of $Y$, longer documents tend to have higher CT scores (while it is not the case for coreference F1 scores).
These obersvations indicates that the conditional distribution $P(Y \mid X_{SpanBERT})$ is very close to the real distribution $P(Y)$ compared to $P(Y \mid X_{GloVe})$ and $P(Y \mid X_{One-hot})$; the mutual information between $X_{SpanBERT}$ and $Y$ is quite large.

\subsection{Genres}
\begin{figure*}[thpb]
\centering
\includegraphics[scale = 0.42]{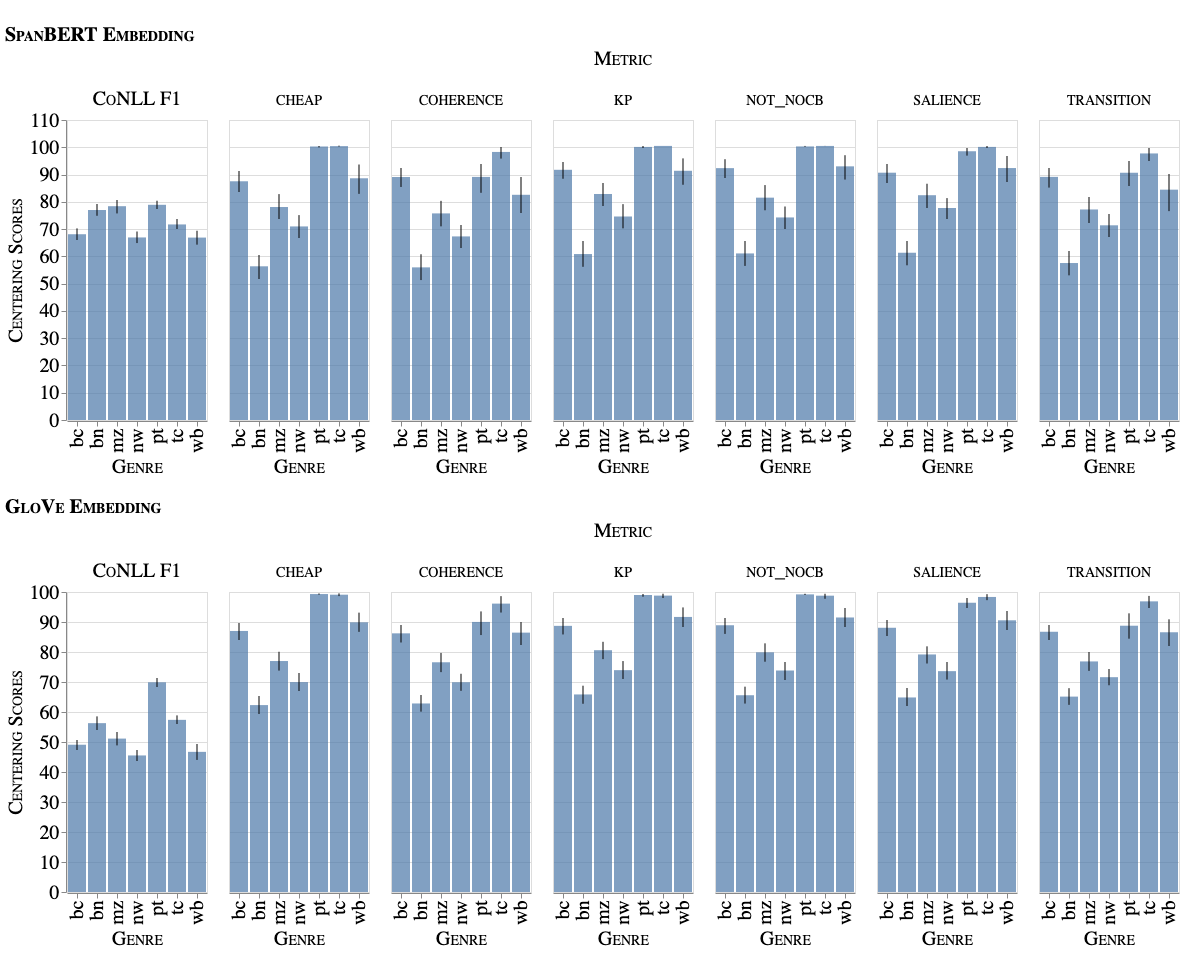}
\caption{The mutual information between coref F1 and CT scores for different genres.}
\label{fig:genres}
\end{figure*}

We report the normalized mutual information and correlations for different genres in Figure \ref{fig:genres}

\end{appendices}

\medskip
\printbibliography